%% file: main.tex
\pdfoutput=1
\documentclass[sigconf,pbalance,nonacm]{acmart}

\usepackage[utf8]{inputenc}
\usepackage{microtype}
\usepackage{graphicx}
\usepackage{xargs}
\usepackage{listings}
\usepackage{booktabs}
\usepackage{multirow}

\usepackage[capitalize]{cleveref}
    \crefname{figure}{Figure}{Figures}
    \Crefname{figure}{Figure}{Figures}
    \crefname{section}{Section}{Sections}
    \Crefname{section}{Section}{Sections}
    \crefname{table}{Table}{Tables}
    \Crefname{table}{Table}{Tables}

\AtBeginDocument{%
  }

\copyrightyear{2025}
\acmYear{2025}
\setcopyright{acmlicensed}
\acmConference[MM '25]{Proceedings of the 33rd ACM International Conference on Multimedia}{October 27--31, 2025}{Dublin, Ireland}
\acmBooktitle{Proceedings of the 33rd ACM International Conference on Multimedia (MM '25), October 27--31, 2025, Dublin, Ireland}
\acmDOI{10.1145/3746027.3755421}
\acmISBN{979-8-4007-2035-2/2025/10}

\begin{document}

\title{Towards Explainable Fake Image Detection with Multi-Modal Large Language Models}

\author{Yikun Ji}
\email{da-kun@sjtu.edu.cn}
\orcid{0000-0001-7788-2361}
\affiliation{
  \institution{Shanghai Jiao Tong University}
  \city{Shanghai}
  \country{China}
}

\author{Yan Hong}
\orcid{0000-0001-6401-0812}
\affiliation{
  \institution{Ant Group}
  \city{Hangzhou}
  \country{China}
}

\author{Jiahui Zhan}
\orcid{0009-0005-6721-6740}
\affiliation{
  \institution{Shanghai Jiao Tong University}
  \city{Shanghai}
  \country{China}
}

\author{Haoxing Chen}
\orcid{0000-0001-6637-8741}
\affiliation{
  \institution{Ant Group}
  \city{Hangzhou}
  \country{China}
}

\author{Jun Lan}
\orcid{0000-0003-0921-0613}
\affiliation{
  \institution{Ant Group}
  \city{Hangzhou}
  \country{China}
}

\author{Huijia Zhu}
\orcid{0009-0008-5784-7225}
\affiliation{
  \institution{Ant Group}
  \city{Hangzhou}
  \country{China}
}
\authornote{Corresponding Authors}

\author{Weiqiang Wang}
\orcid{0000-0002-6159-619X}
\affiliation{
  \institution{Ant Group}
  \city{Hangzhou}
  \country{China}
}

\author{Liqing Zhang}
\orcid{0000-0002-2673-5860}
\affiliation{
  \institution{Shanghai Jiao Tong University}
  \city{Shanghai}
  \country{China}
}
\authornotemark[1]

\author{Jianfu Zhang}
\email{c.sis@sjtu.edu.cn}
\orcid{0000-0002-2673-5860}
\affiliation{
  \institution{Shanghai Jiao Tong University}
  \city{Shanghai}
  \country{China}
}
\authornotemark[1]

\renewcommand{\shortauthors}{Yikun Ji et al.}

\begin{abstract}
  \input{sections/0_abstract}
\end{abstract}




\maketitle

\input{sections/1_intro}
\input{sections/2_related}
\input{sections/3_method}

\input{sections/4_experiment}
\input{sections/5_limitations}
\input{sections/6_conclusion}

\begin{acks}
This research is supported, in part, by the National Natural Science Foundation of China (Grant No. 62302295), by Ant Group, the Shanghai Municipal Science and Technology Major Project, China (Grant No. 2021SHZDZX0102), the Pioneer R\&D Program of Zhejiang Province (No. 2024C01024) and the foundation of Key Laboratory of Artificial Intelligence, Ministry of Education, P.R. China.
\end{acks}

\clearpage
\bibliographystyle{ACM-Reference-Format}
\bibliography{bibs/cherries,bibs/related_work,bibs/semanticscholar_export}

\newpage
\appendix
\input{sections/7_appendix}

\end{document}

%% file: sections/0_abstract.tex
Recent advancements in image generation have provoked social and security concerns, yet most detection methods rely on black-box models that generalize poorly. By utilizing advances in Multi-modal Large Language Models (MLLMs), we propose a framework that fuses six specialized paradigms, each analyzing a distinct aspect of the image, to provide a final verdict with coherent, evidence-based reasoning. Experimental results on a diverse dataset of real and AI-generated images demonstrate that our approach outperforms both traditional detection methods and top humans, while providing explainability. This study underscores the potential of MLLMs in developing robust, explainable, and reasoning-driven detection systems. The code is available at \href{https://github.com/Gennadiyev/mllm-defake}{https://github.com/Gennadiyev/mllm-defake}.

%% file: sections/1_intro.tex
\section{Introduction}
\label{sec:intro}

\begin{figure}[t]
    \centering
    \includegraphics[width=0.46\textwidth]{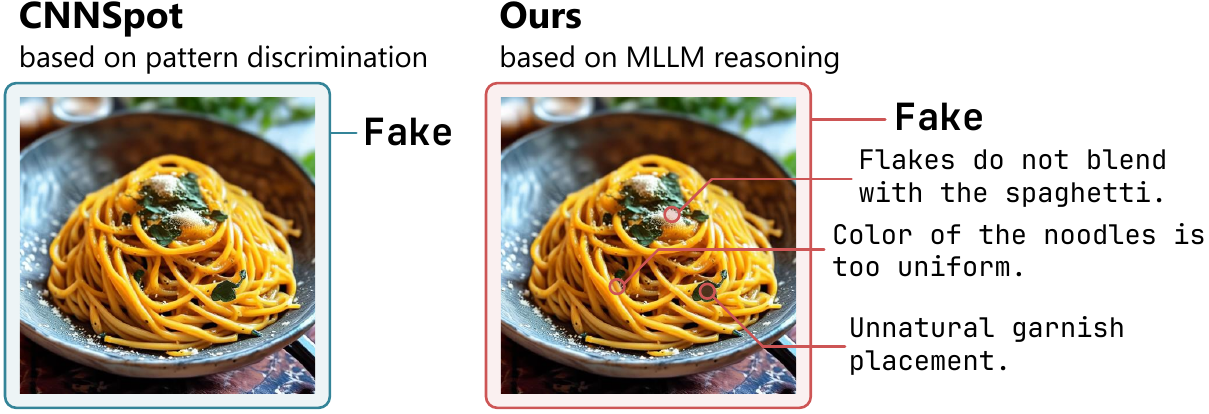}
    \caption{Traditional methods do not provide reasons for their predictions, while our method provides sensible reasons behind the verdict.}
    \Description{An image of a plate of spaghetti, when provided as input to traditional methods, e.g. CNNSpot, we can obtain a verdict that the image is AI-generated. Our MLLM-powered method is able to provide natural language rationale behind the decision.}
    \label{fig:starting}
    \vspace{-16pt}
\end{figure}

\begin{figure*}[t]
    \centering
    \includegraphics[width=\textwidth]{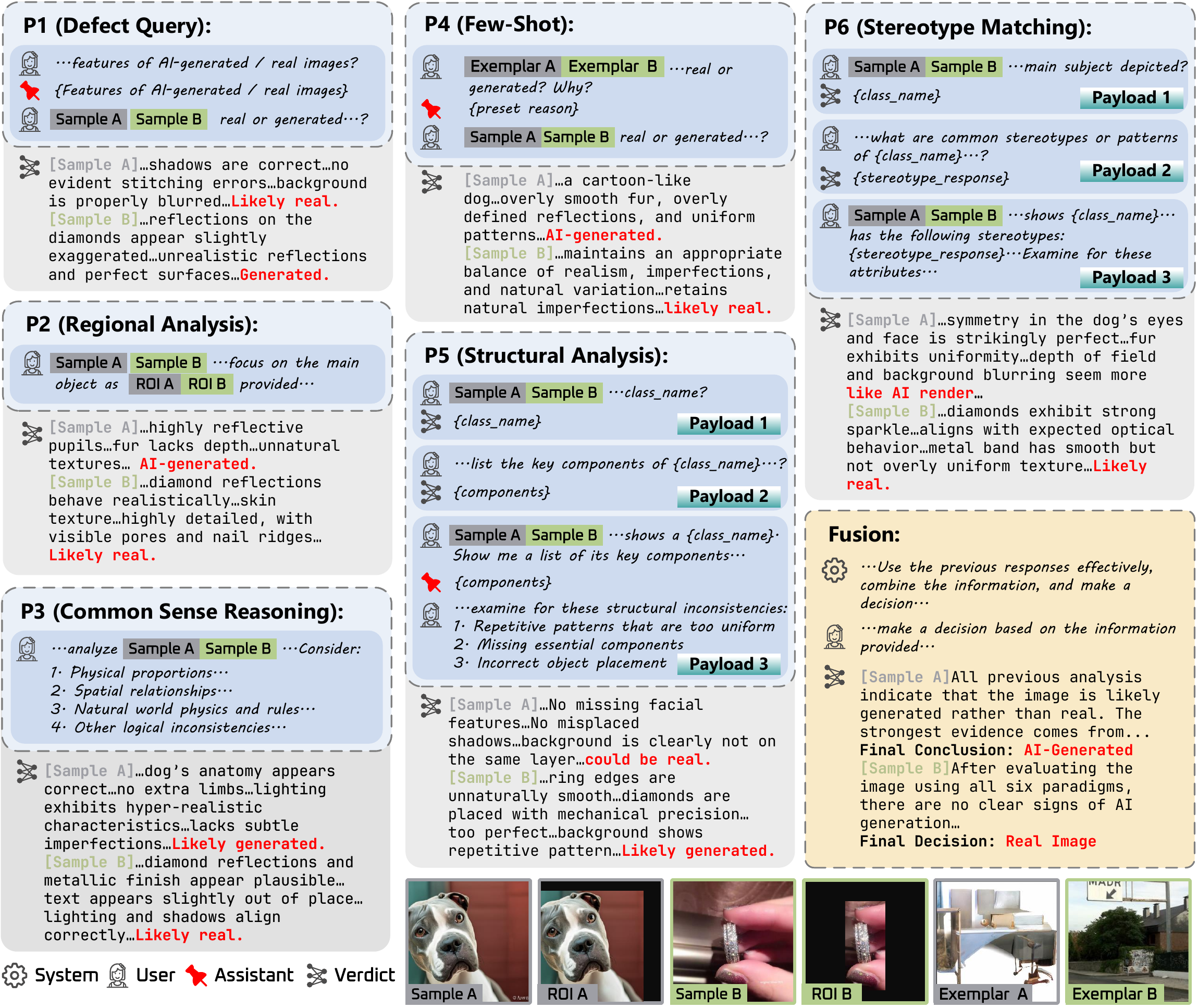}
    \caption{The overall design of the proposed MLLM-based AI-generated image detection framework.}
    \label{fig:pipeline}
    \Description{Sample output from all six prompts and fusion stage. Text is based on two images, an AI-generated dog image, and a natural photograph of a ring.}
    \vspace{-8pt}
\end{figure*}

The rapid emergence of deep learning and artificial intelligence, exemplified by Generative Adversarial Networks (GANs)~\citep{GAN} and diffusion models~\citep{Diffusion}, has revolutionized synthetic image generation, creating remarkable opportunities for creative innovation. Yet, these technologies also pose critical societal challenges, particularly in cyber-security, where the proliferation of synthetic images can threaten both individual and public safety~\citep{SocialIssues}.

Detecting AI-generated images remains a challenging task, usually formulated as a binary classification problem. While deep learning-based methods have exhibited solid performance~\citep{CNNSpot,EfficientNet}, they are prone to overfitting and often rely heavily on large, labeled datasets of real and synthetic images. 
Alternative approaches focus on statistical anomalies~\citep{AEROBLADE, DefakeByReals} but frequently struggle with poor generalization, especially when limited to specific generative models or image domains (\textit{e.g.}, facial images). 
Consequently, there is a pressing need for detection models that not only perform well across diverse generative techniques but also provide interpretability, enabling users to understand why images are flagged.

With the development of multi-modal large language models (MLLMs), researchers have begun exploring their potential in detecting AI-generated images. A key advantage of using MLLMs is their ability to generate predictions with explanations, thus enhancing interpretability.
Studies have shown that an open-source MLLM (mPlug-owl2~\citep{MPlugOwl2}) with 7B parameters can achieve an accuracy of 71.78\%, while OpenAI’s GPT-4V reaches 78.03\%, outperforming the average human accuracy of 74.51\%~\citep{FakeBench}. \citet{ABench} highlights the quality of reasons that MLLMs provide when verifying image authenticity. Although these models demonstrate notable potential, MLLMs remain susceptible to hallucination~\citep{HallucinationLLM} and generally cannot surpass the accuracy of traditional methods. Meanwhile, their responses can be sensitive to prompt design~\citep{VLMPrompting}, suggesting the need for more refined techniques to enhance both accuracy and reasoning quality.

In this work, we enhance the capabilities of multi-modal large language models (MLLMs) through an innovative interrogation methodology. By integrating human expertise with MLLM-based reasoning, we design six distinct prompts, each focusing on a distinct aspect of the image. By fusing the model responses from these prompts, we achieve a higher overall accuracy and improved interpretability.
Our quantitative and qualitative analysis reveals that MLLMs can process visual information similarly to human perception when distinguishing between real and AI-generated images. The models exhibit strong generalization capabilities, with GPT-4o achieving 93.4\% accuracy, outperforming traditional deep learning models~\citep{CNNSpot}~(91.8\%) and AEROBLADE~\citep{AEROBLADE} (85.2\%), as well as the most accurate human annotator (86.3\%). 

This framework highlights the promise of MLLMs for robust, transparent, and explainable AI-generated image detection. Moreover, the model can explicitly articulate why an image is classified as generated, a crucial capability for platforms aiming to provide clear explanations to users, as illustrated in \cref{fig:starting}.

%% file: sections/2_related.tex
\section{Related Works}
\label{sec:related}

\subsection{AI-Generated Image Detection Methods}
AI-generated image detection has received significant attention due to the rapidly improving fidelity of synthetic images produced by Generative Adversarial Networks (GANs)~\citep{GAN}, autoregressive transformers\citep{VQVAE}, and diffusion-based models~\citep{SD21,SD3,DDIM,DDPM}. Early research primarily focused on artifact detection, extracting discriminative cues from spatial and frequency domains to highlight inconsistencies between real and synthetic images~\citep{ArtifactDetSimGAN, ArtifactUseFreq, ArtifactWhatsDetectable, ArtifactSpectrumDefake, ArtifactTextureDefake}. These inconsistencies often stem from upsampling artifacts in GAN pipelines~\citep{ArtifactUseFreq}, texture synthesis mismatches\citep{ArtifactTextureDefake}, or limited high-frequency decay in generated outputs\citep{ArtifactSpectrumDefake}. By learning to detect these subtle traces, models can effectively expose fakes, even when artifacts are visually imperceptible.

Beyond artifact-based methods, deep-learning based methods are also advancing. ResNet and Vision Transformers trained on real and synthetic samples can effectively distinguish real from AI-generated images~\citep{CNNSpot, EfficientNet, ComFor}. This paradigm leverages deep networks' strong feature extraction capabilities to automatically learn discriminative patterns. However, the effectiveness of large-scale training heavily depends on data diversity and quality. Generalization to unseen generative models, particularly from GANs to diffusion-based methods, remains challenging. Meanwhile, as diffusion models produce increasingly photorealistic images, traditional artifact-based cues, including unnatural hues, faulty perspectives, and blurry text, are no longer reliable. Additionally, \citet{FakeOrJPEG} highlighted that existing models are prone to overfitting on JPEG artifacts, further limiting their robustness.

One branch of complementary research direction focuses on improving model explainability. Most existing detection methods output only a binary classification, failing to indicate how or where synthetic cues are identified. Recent efforts emphasize fine-grained or localized detection, providing more transparent classification rationales. Notable approaches include measuring reconstruction errors between an input image and its diffusion-based reconstruction~\citep{AEROBLADE}, leveraging multi-branch systems to produce multi-level labels for generated images~\citep{DefakeByReals}, or computing local intrinsic dimensionalities~\citep{IntrinsicDimensionalities}. Some approaches integrate text-image contrastive learning for better explainability~\citep{XplainedAxiomaticAttr,XplainedCNN,XplainedGradCAM}.

While these methods provide regional clues, they lack human-understandable explanations, limiting their effectiveness in enhancing explainability.

\subsection{Multi-Modal Large Language Models}
Multi-modal large language models (MLLMs) are large language models (LLMs) enhanced with vision or audio modalities~\citep{VILA,Fuyu8B}. Over the past few years, both open-source and proprietary models have significantly advanced vision-language capabilities. Open-source ones include Qwen-VL~\citep{QwenVL,Qwen2VL,Qwen25VL} and QVQ~\citep{QVQ}, InternVL~\citep{InternVL}, Ovis~\citep{Ovis}, LLaVA and its predecessors~\citep{LLaVA, ImprovedLLaVA, LLaVANeXT}, Llama-3~\citep{Llama3}, among many others. Proprietary models have also progressed rapidly. Notable releases include GPT-4o, GPT-4o-mini~\citep{GPT4o,GPT4oMini} and Gemini-1.5~\citep{Gemini,Gemini15}. These models demonstrate remarkable performance in vision-related tasks, including image reasoning and interpretation.
Most benchmarks for vision-language models focus on reasoning capabilities related to the image~\citep{MMTBench,QSpatial} or geometric problem-solving skills~\citep{VisOnlyQA,MathVista}.

\citet{FakeBench} proposed FakeBench, a benchmark for general AI-generated image detection that includes three primary datasets: reasoning, interpretation, and open-question answering. It covers ten generative models across diverse image content, revealing that current MLLMs demonstrate moderate detection accuracy, basic reasoning abilities, and limited fine-grained authenticity analysis. \citet{ABench} created A-Bench, which ``challenges both high-level semantic understanding and low-level visual quality perception'' of MLLMs in the form of multiple-choice questions, pointing out major flaws in each AI-generated image. The question-answer pairs are curated by humans and not actively generated by MLLMs. FakeReasoning~\citep{FakeReasoning} conducts the forgery reasoning task in structured steps using MLLMs, giving promising results by fine-tuning LLaVA-1.5-13b on the MMFR-Dataset, created using GPT-4o and human experts.

Our proposed method significantly enhances the accuracy while also providing fine-grained explanations of why an image is considered real or generated, addressing these limitations.

%% file: sections/3_method.tex
\section{Methodologies}
\label{sec:method}
Given an image $x$, we employ a detection function $f$ to classify it as either real (\textit{i.e.}, natural) or fake (\textit{i.e.}, AI-generated), producing a prediction $y = f(x)$.
Our approach is driven by the observation that different prompt formulations enable MLLMs to focus on distinct aspects of an image.

Techniques such as image cropping~\citep{DetailPerceptionInZeroShotVQA}, few-shot learning~\citep{MMIContextL,VLFewShotExamples}, and region-of-interest selection~\citep{VStar} enhance MLLM performance by refining visual feature extraction and interpretation. These techniques influence how models extract and analyze relevant image attributes, leading to more effective detection.
Prompt formulation plays a critical role in MLLM performance. We design optimized prompts that effectively integrate textual and visual information, mimicking human perception. By combining domain expertise with MLLM reasoning, our approach enhances both explainability and accuracy. The proposed framework is illustrated in \cref{fig:pipeline}, and the following sections provide a detailed explanation of our prompt construction and detection pipeline.

\subsection{Prompt Engineering}
\label{subsec:prompt}
We provide visual images and textual prompts to the MLLMs, leveraging chain-of-thought (CoT) prompting to improve reasoning.
Our approach structures MLLM \textit{sessions}~(\textit{i.e.}, context window) as sequential queries, where each query is defined as:
$o_i=M(t_i,v_i,c_{i-1})$, where $i$ denotes the current query, $M$ represents the MLLM, $o$ is the model’s output, $t$ and $v$ are text/visual prompts (image type in this task) from the user, and $c$ is the session context (query history) from previous queries.
Each query triggers a computational \textit{payload}, incurring processing costs.
The output $o$ is incorporated into the next query as part of the updated context: $c_{i+1} = (t_i, v_i, o_i, c_i)$.
To ensure consistency and mitigate randomness, we utilize a predefined $o$ as an \textit{assistant} reference, summarized by the user. This approach also optimizes inference efficiency by reusing previous outputs, improving model stability and reducing computational costs while maintaining fast inference speed.

\paragraph{Verdict Prompt (P0).}
In our method, a designated \textit{verdict prompt} $n$ is introduced, where the input image $v_n$ corresponds to $x$, and the prompt $t_n$ is structured as follows: 
\par
\texttt{[Is this image real or fake? End your response with either "real" or "fake".]} 
\par
This enforces a concise model prediction, allowing us to directly use the output $o_n$ as the final classification result $y$. 
P0 serves as the baseline of our framework, leveraging MLLMs' classification capabilities. Unless otherwise specified, our approach always concludes with P0.

To systematically explore prompt variations, we develop six distinct prompt strategies, categorized into three types: General Prompts (direct analysis mimicking human reasoning), Few-Shot Prompts (providing a few examples to guide model predictions) and Content-Based Prompts (first analyzing image content before predicting).
We briefly introduce each prompt type later. For full implementation details, please refer to the provided code.

\paragraph{System Prompt.}
To enhance the interoperability of our method, we define an initial system prompt $h_0$ that defines the session's general settings:
\par
\texttt{[You are an AIGC detection specialist. The user will ask you about whether an image is real or generated. Observe the image carefully to decide whether it is real or generated. Explain your reasons, and end your response with either "real" or "generated".]}
\par
This ensures that the model justifies its classification decision, maintaining explainability in the prediction function $f$.
For all subsequent prompts P1–P6, this system prompt initializes the session.

\subsection{General Prompts}
Humans analyze images top-down, first grasping the overall scene before focusing on key objects and details. We design our model to follow a similar process, leveraging MLLMs' reasoning abilities to detect anomalies.
Our framework systematically analyzes primary objects, assessing their properties against real-world constraints and logical coherence. 
We design three prompts: Defect Query (P1): Conducts a broad inspection to identify general anomalies; (P2): Focuses on specific regions to refine local anomaly detection; and Common Sense Reasoning (P3): Applies logical reasoning to verify plausibility against real-world knowledge.
By integrating these perspectives, our approach enhances the reliability and explainability of AI-generated image detection.

\paragraph{Defect Query (P1).} 
Generative models rely on extensive real-world data, yet no dataset can fully encapsulate all causal factors of a scene. Photography follows a cause-and-effect paradigm—objects are shaped by lighting, camera angles, lens parameters, and the photographer’s intent. Models trained solely on photographic data often fail to internalize these relationships, leading to artifacts such as incorrect shadows, unnatural object placements, and abnormal focus or blur.
To address this, the defect query prompt directs the MLLM to identify such anomalies. The model is informed that the image may contain defects and is tasked with listing potential issues. 
The process consists of three queries. In the first two queries, $o_1$ lists known AI-generated artifacts, $o_2$ highlights features typical of real images, and $t_1$ / $t_2$ are
\par \texttt{[What are common defects in AI-generated images?]} / \texttt{[What are the features that real images often have?]} \par
Finally the model makes a final prediction with P0, reasoning based on $o_1$ and $o_2$.
This query structure enables MLLMs to predict and justify their classification by identifying defects such as abnormal object proportions, unrealistic reflections, etc.\looseness=-1

\paragraph{Regional Analysis (P2).} 
Each input image contributes a fixed number of tokens to the MLLM. To emphasize key areas, this method extracts Regions of Interest (ROI) from the original image, allowing the model to focus on specific objects or sub-scenes, such as facial features or text on a book. 
While this approach may omit broader context, it enhances detail analysis in a controlled setting—especially useful when the main object occupies only a small portion of the frame. 
This method is inspired by prior work on synthetic image detection~\citet{XplainedCNN}, where heatmaps are used to improve model explainability.
The process consists of two queries. In the first query, $v_1 = x$, $o_1$ will request user confirmation of the ROI, and $t_1$ is
\par
 \texttt{[Instead of analyzing the full image, focus on the ROI in the next image to determine whether it is real or generated.]}
\par
We extract the top three most relevant regions using DINOv2~\cite{DinoV2}, generating heatmaps that are thresholded to form bounding boxes from binary pixel maps. Pretrained weights from ImageNet are used.
For the second query, $v_2$ is the ROI image and $t_2$ is
\par
\texttt{[Here is the ROI. Please indicate whether the image is real or generated based on ROI. End your response with either "real" or "generated"]}. 
\par
The final prediction is then generated, along with an explanation focusing on the main area of the image. This ensures that the model prioritizes fine details of key objects, such as pattern distortions, unrealistic textures, and other detail problems, enhancing detection accuracy.

\paragraph{Common Sense Reasoning (P3).} 
AI-generated images often contain logical inconsistencies that are implausible in the real world, such as incorrect numbers of fingers, unnatural object attachments, and distorted text. While humans easily recognize these anomalies, MLLMs may overlook them when prioritizing overall image realism.
To address this, and inspired by~\cite{CommonSenseReasoning}, we prompt the MLLM to detect inconsistencies by comparing images against real-world patterns.
This method uses a single query with $v_1 = x$ and $t_1$ is 
\par 
\texttt{[
Please analyze this image for any violations of common sense or real-world logic. 
Consider: 
1. Physical proportions (e.g., number of fingers, limbs);
2. Spatial relationships between objects, such as merged textures; 
3. Natural world physics and rules, such as unnaturally bent fences; 
4. Other logical inconsistencies in the scene Explain any inconsistencies you find. 
If you find many unexplainable inconsistencies, the image is likely generated.]} 
\par
This approach guides the model to examine critical visual cues, such as limb counts and spatial relationships. While these anomalies may be subtle, they serve as strong indicators of AI-generated content. By incorporating logical consistency checks, this method improves model generalization across diverse images while enhancing interpretability in ``generated'' classification.\looseness=-1

\subsection{Few-Shot Prompt}

\paragraph{Few-Shot (P4).} 
In the realm of MLLM, few-shot learning refers to the technique of providing exemplars in the prompt before the query. Our few-shot prompt provides two labeled exemplars along with their human-annotated responses to a baseline prompt, refining the model’s conceptual understanding. 
Our prompt design consists of three turns, each using a structured text prompt similar to P0 for prediction. The first two queries introduce labeled exemplars, one real for $v_1$ and one fake for $v_2$, with manually annotated justifications for their classification. These responses replace $o_1$ and $o_2$, and are packaged with previous images in a single history payload. Finally, P0 will be conducted for the final prediction.
This method leverages the MLLM’s reasoning and generalization capabilities under data constraints, ensuring interpretability. By referencing only a few exemplars, the model’s decision process remains transparent and grounded in direct comparisons. The compact dataset encourages the model to focus on essential features, such as object outlines, textures, and logical color consistency, instead of relying on memorized distributions, leading to more robust conclusions.

\subsection{Content-Based Prompts}
Content-Based Prompts evaluate an image’s internal coherence by analyzing its main subject, structural patterns, and alignment with real-world expectations. These prompts leverage the MLLM’s ability to identify key features and provide interpretable justifications for inconsistencies.
To prevent conflicts between different queries, the process follows a structured multi-session approach.
The first session is Object Identification, where 
$t$ is 
\par
\texttt{[What is the main subject in the image? Please answer concisely with a short descriptive noun phrase, e.g., 'a red bus', 'a daisy'.]}
\par
The second session then builds upon the response from the first turn, refining the analysis with $v=x$. $o$ will be the \{class\} of $x$.
Then the \{class\} information is integrated into the following sessions for the final prediction, which will be detailed in the next paragraphs.

\paragraph{Structural Analysis (P5).} 

This method systematically identifies internal inconsistencies within an image based on its object class. The detection process follows a two-step query approach:
1. Component Identification, where
$t$ is
\par
\texttt{[This image shows a \{class\}. List its key components before determining whether it is real or generated.]}
\par
$v=x$, and $o$ outputs a list of expected components based on the identified object class.
2. Structural Verification: The model checks for missing or misplaced components, assessing whether these structural faults indicate AI generation. At the same time, P0 is applied to finalize the classification.
This modular approach enhances interpretability by mapping each detected anomaly to a logical explanation, conditioned on the expected class structure. Using this prompt, the model can effectively detect AI-generated images with: unnaturally uniform repetitions (\textit{e.g.}, identical windows in a building, consistent distortions across a fence); absence of critical features (\textit{e.g.}, a dog without a tail, a person missing hands); implausible positioning of objects (\textit{e.g.}, a tree growing out of a building, a car parked on a rooftop).

\paragraph{Stereotype Matching (P6).} 
This method first classifies the dominant object in an image, then evaluates whether its features conform too closely or uniformly to common stereotypes. Examples include excessively symmetrical human faces, unnaturally uniform animal textures, or exaggerated stereotypical attributes (\textit{e.g.}, a red-beaked rooster with atypical pigmentation).
The detection will first conduct stereotype analyses session, where
$t$ is 
\par
\texttt{[If an image shows \{class\}, what are some common stereotypes or patterns you would expect? List at least three.]}
\par
$o$ outputs common stereotypes of the identified class.
Then, we start a new session. The model analyzes the image for the presence of these common stereotypes, explaining any anomalies and determining whether the image is real or AI-generated with P0.
A certain degree of stereotypical accuracy enhances realism, but excessive adherence to these patterns may indicate AI generation, reflecting biases learned from training data. By prompting the MLLM to assess how closely an image follows stereotypical features, this method enhances interpretability in detecting unnatural uniformity or exaggerated regularities that deviate from real-world variation.

\subsection{Fusion Process}

Prompts categorized as \emph{General}, \emph{Few-Shot}, and \emph{Content-Based} enhance AI-generated image detection via MLLMs by targeting interpretability aspects like local inspections (P2, P4), logical consistency (P3, P5), and real-world traits (P1, P6).
We propose two fusion methods to determine the final verdict.
The primary approach executes P1–P6 in parallel before consolidating an output, balancing accuracy and explainability.
Alternatively, a majority vote method tallies individual prompt verdicts, trading some explainability for computational efficiency while remaining effective in most cases. Experimental results validate these trade-offs, showing the sequential method excels in interpretability and the majority vote offers practical efficiency.

%% file: sections/4_experiment.tex
\section{Experiments}
\label{sec:experiments}

\subsection{Experimental Setup}

\input{figures/result}

We ensure relevance and effectiveness in real-world AI-generated image detection by covering a diverse range of generation models, including but not limited to GANs, auto-regressive models, and diffusion models. To achieve this, we selected 1,000 real images and 1,000 AI-generated images from WildFake \citep{WildFake}, spanning different generation architectures, with detailed distribution provided in the \textit{appendix}.
To benchmark MLLM performance against humans, we invited 24 volunteers to classify the images, 12 of whom had prior experience with AI image generation.

We tested recent MLLMs such as GPT-4o, GPT-4o-mini, Llama 3.2 VI (Llama-3.2-Vision-Instruct 11B)~\citep{Llama3}, LLaVA-CoT~\citep{LLaVACoT}, Qwen-VL~\citep{QwenVL, Qwen2VL, Qwen25VL, QVQ}, Ovis~\citep{Ovis} and InternVL series~\citep{InternVL}. Among them, we selected four models for our full analyses: GPT-4o, GPT-4o-mini, Llama 3.2 VI, and LLaVA-CoT.
In our experiments, we used OpenAI GPT-4o that points to \verb|gpt-4o-| \verb|2024-08-06| and GPT-4o-mini that points to \verb|gpt-4o-mini-| \verb|2024-07-18|. The Llama 3.2 VI refers to the Llama-3.2-Vision-Instruct 11B model, from which LLaVA-CoT is a fine-tuned chain-of-thought model. Local MLLMs are deployed using \verb|vllm-0.7.2|~\citep{vllm} on four NVIDIA A100-40G GPUs.

To compare the performance with traditional methods, we select AEROBLADE~\citep{AEROBLADE}, CNNSpot~\citep{CNNSpot}, CommunityForensics~\citep{ComFor} and ObjectFormer~\citep{ObjectFormer}. We evaluate them on the same sampled dataset used for MLLM and human evaluations.
For AEROBLADE, we employ Stable Diffusion versions 1.5, 2.1, and 3.5-large~\citep{SD21, SD3} as reconstruction models and select SD3.5-Large for the results, since it provides the best performance.
For CNNSpot and ObjectFormer, since the pretrained models performed suboptimally, we evaluated on checkpoints that we trained on the same data.
CommunityForensics requires no training.

\begin{figure*}[t]
    \centering
    \includegraphics[width=\linewidth]{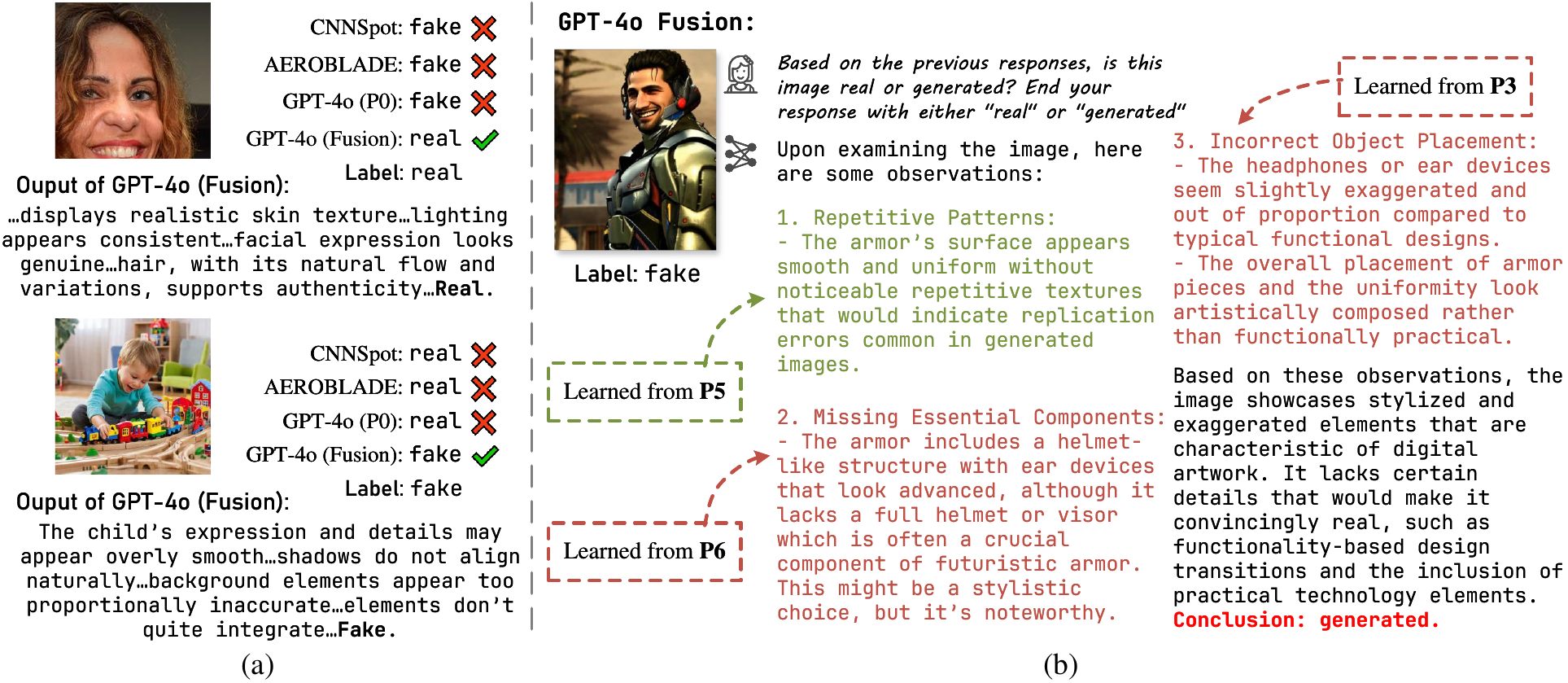}
    \caption{(a) Examples where GPT-4o fusion gives correct result, while CNNSpot, AEROBLADE, and vanilla GPT-4o (P0) fail. (b) A fusion example. GPT-4o can combine responses from P1-6 effectively, drawing conclusions from \emph{reasons} instead of verdicts.}
    \Description{A detailed figure showing two parts: (a) examples of GPT-4o fusion correctly identifying cases where CNNSpot, AEROBLADE, and vanilla GPT-4o fail; and (b) excerpt of text generated from the fusion process where GPT-4o combines responses from multiple prompts P1-6, focusing on reasoning rather than just verdicts.}
    \label{fig:good_cases}
\end{figure*}

\input{figures/appendices/timeit}

\cref{tab:timeit} reports the average inference time per image when six prompts are queried both sequentially and with parallelism enabled.
For Llama 3.2 VI~\citep{Llama3} and LLaVA-CoT~\citep{LLaVACoT}, profiling was conducted on 8× NVIDIA A100-40G GPUs. Network latency is included for models relying on OpenAI API calls~\citep{GPT4o,GPT4oMini}.
Results indicate that P5 and P6 require more time, as they involve larger, multi-turn payloads, but all models maintain reasonable inference times with parallelism, given that natural language output remains rate-limited by token throughput of MLLMs.

\subsection{AI-Generated Image Detection Accuracy}

The results are summarized in \cref{tab:result}, reporting accuracy for MLLMs: GPT-4o, GPT-4o-mini, Llama 3.2 VI, LLaVa-CoT; traditional methods: AEROBLADE~\citep{AEROBLADE}, CNNSpot~\citep{CNNSpot}, CommunityForensics~\citep{ComFor}, DMimageDetection~\citep{DMImageDetection} and NPR~\citep{NPR}; and human performance.
For MLLMs, we report results for baseline (P0), all proposed prompts, majority vote of P1-P6 (Maj.), and fused results.
Humans performed well overall (81.9\% accuracy), particularly on real images (84.8\% accuracy). The best volunteer achieved 4.4\% higher accuracy than the average human performance and slightly outperformed the best MLLM baseline (86.3\% vs. 85.2\% for GPT-4o P0).
Other MLLMs performed worse than GPT-4o. 
While CNNSpot (91.8\%) surpassed GPT-4o (85.2\%), it operates as a black-box model.

Not all P1–P6 outperform P0, which is expected since these prompts prioritize both accuracy and interoperability.
This is a trade-off rather than a direct improvement in both aspects.
As a scaled-down version of GPT-4o, GPT-4o-mini performs significantly worse across prompts. 
They have different trends with different prompts.
GPT-4o excels in P3 (reasoning) and P6 (stereotype matching). 
GPT-4o-mini struggles the most on P3, while in P1 (defect detection), GPT-4o outperforms GPT-4o-mini, particularly in detecting generated images.

A key finding is the performance gain from response fusion. When using majority vote from verdicts provided by P1~P6, we consistently achieve better accuracy over independent prompts. To fully utilize previous responses, when fusing together for a final verdict, GPT-4o and Llama 3.2 Vision Instruct show an increase in accuracy. While the other two models did not benefit from this process, the aggregated reasons provided by the fusion process are more sensible than individual outputs. When fused, GPT-4o can reach 93.4\% accuracy, surpassing CNNSpot.
Both majority voting and fusion give more accurate results when compared to P0: \textit{Fused compared to P0, GPT-4o +8.2, GPT-4o-mini + 10.7, Llama 3.2 VI + 7.8, LLaVA-CoT + 10.4.}

These results align with prior research, demonstrating that model ensembling enhances generalization, accuracy, and reduces false positives and false negatives in MLLM-based AI-generated image detection.

\input{figures/generated_or_fake}

\subsection{Prompt Effectiveness}
\label{subset:ablation}

\begin{figure}[htbp]
    \centering
    \includegraphics[width=0.9\linewidth]{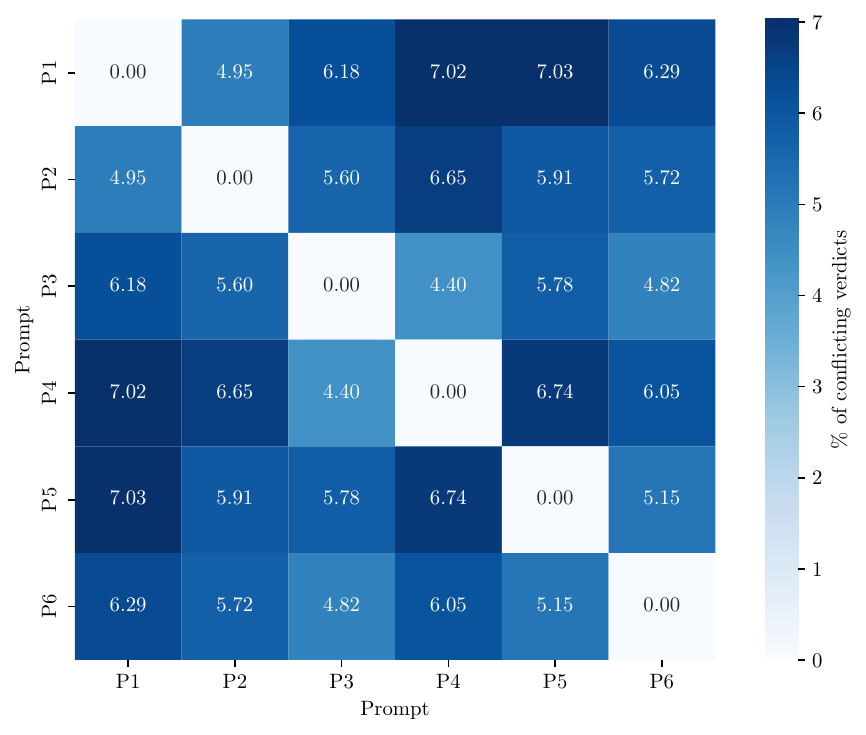}
    \caption{Percentage of cases where prompts give different verdicts. The data is aggregated from all four models evaluated.}
    \Description{A heatmap showing a verdict mismatch between prompts. The smallest mismatch is 4.40\% from P3 and P4; the largest mismatch is 7.03\% from P1 and P5.}
    \label{fig:prompt_conflicts}
\end{figure}

To assess the effectiveness of our six prompts, we calculated the probability of divergent verdicts across the prompts for a given image, with results presented in~\cref{fig:prompt_conflicts}. Among all real and generated samples, 22.31\% received at least one contradictory verdict across prompts from GPT-4o, and 31.44\% from LLaVA-CoT. This demonstrates that different prompts can guide the model to focus on different aspects of the image, leading to independent, different decisions.

We also conducted an ablation study by systematically excluding each prompt from the fusion process while preserving the reasoning structure. By analyzing the resulting accuracy, we assess the significance of each prompt in enhancing the model's decision-making process.
If a prompt is redundant or ineffective, its removal should have minimal impact on the fused accuracy. Conversely, if a prompt is critical, omitting it should lead to a noticeable accuracy drop.
For each iteration, we remove one prompt at a time and re-run the fusion queries, comparing the results against the original fused accuracy reported in ~\cref{tab:result}.
Different models exhibited different sensitivities to prompt removal. For example, GPT-4o and GPT-4o-mini showed the least accuracy drop when omitting P1 (defect detection), whereas open-source models retained more accuracy when ablating P1. 
No ablation configuration outperformed the full fusion model (P1–P6), demonstrating the effectiveness of all proposed prompts.

\subsection{Qualitative Results on Reasoning}

\cref{fig:good_cases} (a) presents two cases where traditional methods and P0 fail, but fused GPT-4o correctly classifies the images by integrating regional observations and general cues.
In the first case, GPT-4o with fused prompts identifies realistic skin texture, consistent lighting, genuine facial expression, and naturally flowing hair, leading to a real classification.
In the second case, GPT-4o with fused prompts detects overly smooth facial details, unnatural shadows, disproportionate background elements, and poor integration, resulting in a fake classification.
These cases demonstrate GPT-4o’s holistic reasoning, effectively combining multiple perspectives for accurate detection.

A key advantage of our pipeline is the MLLM fusion stage, which corrects errors from individual prompts (P1–P6). 
As shown in \cref{fig:good_cases} (b), while P5 incorrectly classifies an image as real, the model re-evaluates observations from P3 and P6, concluding that the image is AI-generated.
By integrating previous responses, the model enhances explainability, reinforcing final classifications with well-structured reasoning. 

\subsection{Discussions on MLLM Rejections}

MLLMs may generate non-parsable responses or refuse to comment on the given image, termed ``rejections,'' when processing certain images. The rejection rate varies across different models, and the types of images rejected differ accordingly. A frequent justification for these rejections by MLLMs is their inability to provide commentary on images, particularly those depicting faces, due to ethical considerations.
These rejections can be mitigated by modifying the prompt, leading to lower rejection rates.
\Cref{tab:generated_or_fake} shows that among the 2,000 images evaluated, replacing the word ``fake'' with ``generated'' resulted in a 0.95\% (19 samples) reduction in rejections for GPT-4o and 1.3\% for GPT-4o-mini. In comparison, open-source models generally exhibit lower rejection rates. Open-source models such as Llama-3.2-Vision-Instruct and LLaVA-CoT have a much lower rejection rate than GPT-4o and -4o-mini.

%% file: figures/result.tex
\begin{table*}[htbp]
\centering
\resizebox{\textwidth}{!}{ 
\renewcommand{\arraystretch}{1.00}

\begin{tabular}{@{}ccccccccccccc@{}}
\toprule
\multirow{2}{*}{Prompt} & \multicolumn{3}{c}{\textbf{GPT-4o}}                              & \multicolumn{3}{c}{\textbf{GPT-4o-mini}}                         & \multicolumn{3}{c}{\textbf{Llama 3.2 VI}}                        & \multicolumn{3}{c}{\textbf{LLaVA-CoT}}           \\ \cmidrule(l){2-13} 
                        & All            & Real           & \multicolumn{1}{c|}{Generated} & All            & Real           & \multicolumn{1}{c|}{Generated} & All            & Real           & \multicolumn{1}{c|}{Generated} & All            & Real           & Generated      \\ \midrule
P0                      & 0.852          & 0.855          & 0.849                          & 0.769          & 0.910          & 0.628                          & 0.561          & 0.645          & 0.477                          & 0.594          & 0.768          & 0.420          \\ \midrule
P1                      & 0.846          & 0.802          & 0.888                          & 0.776          & 0.972          & 0.579                          & 0.561          & 0.645          & 0.477                          & 0.592          & 0.749          & 0.435          \\
P2                      & 0.848          & 0.824          & 0.871                          & 0.776          & 0.846          & 0.706                          & \textbf{0.595}    & 0.637          & 0.552                          & -              & -              & -              \\
P3                      & \textbf{0.896}    & 0.922          & 0.869                          & 0.701          & 0.721          & 0.680                          & 0.531          & 0.412          & 0.650                          & 0.600          & 0.619          & 0.580          \\
P4                      & 0.834          & 0.801          & 0.867                          & \textbf{0.860}    & 0.874          & 0.846                          & 0.497          & 0.525          & 0.468                          & 0.526          & 0.583          & 0.469          \\
P5                      & 0.841          & 0.813          & 0.868                          & 0.734          & 0.732          & 0.735                          & 0.497          & 0.297          & 0.697                          & \textbf{0.610}    & 0.752          & 0.467          \\
P6                      & 0.888          & 0.880          & 0.895                          & 0.753          & 0.767          & 0.738                          & 0.584          & 0.563          & 0.604                          & 0.586          & 0.603          & 0.568          \\ \midrule
w/o P1                  & 0.895          & 0.917          & 0.872                          & \textbf{0.829}    & 0.862          & 0.796                          & 0.605          & 0.512          & 0.698                          & 0.669          & 0.685          & 0.653          \\
w/o P2                  & 0.863          & 0.850          & 0.876                          & 0.790          & 0.823          & 0.757                          & 0.609          & 0.534          & 0.683                          & 0.674          & 0.719          & 0.628          \\
w/o P3                  & 0.871          & 0.850          & 0.891                          & 0.799          & 0.835          & 0.763                          & \textbf{0.628}    & 0.603          & 0.653                          & 0.667          & 0.706          & 0.627          \\
w/o P4                  & \textbf{0.913}    & 0.923          & 0.903                          & 0.770          & 0.851          & 0.689                          & 0.612          & 0.523          & 0.700                          & \textbf{0.690}    & 0.725          & 0.654          \\
w/o P5                  & 0.865          & 0.848          & 0.882                          & 0.786          & 0.856          & 0.715                          & 0.623          & 0.580          & 0.666                          & 0.665          & 0.683          & 0.647          \\
w/o P6                  & 0.875          & 0.874          & 0.875                          & 0.765          & 0.826          & 0.704                          & 0.587          & 0.515          & 0.658                          & 0.677          & 0.720          & 0.633          \\ \midrule
\textbf{Maj.}           & 0.925          & 0.949          & 0.901                          & \textbf{0.878} & \textbf{0.895} & \textbf{0.861}                 & 0.634          & 0.653          & 0.615                          & \textbf{0.697} & \textbf{0.769} & \textbf{0.625} \\ \midrule
\textbf{Fusion}         & \textbf{0.934} & \textbf{0.955} & \textbf{0.912}                 & 0.876          & 0.893          & 0.859                          & \textbf{0.639} & \textbf{0.660} & \textbf{0.617}                 & 0.696          & 0.767          & \textbf{0.625} \\ \midrule
\multirow{2}{*}{}       & \multicolumn{3}{c}{\textbf{AEROBLADE}}                           & \multicolumn{3}{c}{\textbf{CNNSpot}}                             & \multicolumn{3}{c}{\textbf{CommunityForensics}}                  & \multicolumn{3}{c}{\textbf{ObjectFormer}}        \\ \cmidrule(l){2-13} 
                        & All            & Real           & \multicolumn{1}{c|}{Generated} & All            & Real           & \multicolumn{1}{c|}{Generated} & All            & Real           & \multicolumn{1}{c|}{Generated} & All            & Real           & Generated      \\ \midrule
-                       & 0.852          & 0.834          & 0.870                          & \textbf{0.918} & \textbf{0.978} & \textbf{0.857}                 & 0.861          & 0.850          & 0.872                          & 0.903          & 0.910          & 0.815          \\ \midrule
\multirow{2}{*}{}       & \multicolumn{3}{c}{\textbf{DMimageDetection}}                    & \multicolumn{3}{c}{\textbf{NPR}}                                 & \multicolumn{3}{c}{\textbf{Human (Average)}}                     & \multicolumn{3}{c}{\textbf{Human (Best)}}        \\ \cmidrule(l){2-13} 
                        & Public         & Real           & \multicolumn{1}{c|}{Generated} & All            & Real           & \multicolumn{1}{c|}{Generated} & All            & Real           & \multicolumn{1}{c|}{Generated} & All            & Real           & Generated      \\ \midrule
-                       & 0.871          & 0.924          & 0.870                          & 0.787          & 0.915          & 0.660                          & 0.819          & 0.848          & 0.790                          & \textbf{0.863} & \textbf{0.910} & \textbf{0.815} \\ \bottomrule
\end{tabular}

}
\caption{Quantitative Comparison Results on Accuracy.
The fusion of P1-6 is giving the best result across all MLLMs, with GPT-4o surpassing traditional methods and the best human on the task. The Maj. row shows the majority vote result.}
\label{tab:result}
\end{table*}

%% file: figures/appendices/timeit.tex
\begin{table}[t]
\centering
\resizebox{\columnwidth}{!}{%
\renewcommand{\arraystretch}{1.05}

\begin{tabular}{@{}ccccc@{}}
\toprule
\textbf{Time (s.)}   & \textbf{GPT-4o} & \textbf{GPT-4o-mini} & \textbf{Llama 3.2 VI} & \textbf{LLaVA-CoT} \\ \midrule
P0                   & 1.4             & 0.9                  & 3.2                   & 3.4                \\
P1-6                 & 47.9            & 25.0                 & 67.7                  & 69.1               \\
P1-6 + F.            & \textbf{52.2}   & \textbf{29.6}        & \textbf{78.1}         & \textbf{81.5}      \\
P1-6 (Parellel)      & 11.4            & 7.2                  & 32.8                  & 34.0               \\
P1-6 (Parellel) + F. & 16.2            & 6.8                  & 37.9                  & 41.3               \\ \bottomrule
\end{tabular}

}
\caption{Evaluation time for different models and different prompts, in seconds per image. The time reported includes image processing.}
\label{tab:timeit}
\end{table}

%% file: figures/generated_or_fake.tex
\begin{table*}[t]
\centering
\renewcommand{\arraystretch}{1.05}
\begin{tabular}{@{}ccccccccc@{}}
\toprule
\multirow{2}{*}{Prompt} & \multicolumn{2}{c}{\textbf{GPT-4o}}  & \multicolumn{2}{c}{\textbf{GPT-4o-mini}} & \multicolumn{2}{c}{\textbf{Llama 3.2 VI}} & \multicolumn{2}{c}{\textbf{LLaVA-CoT}} \\ \cmidrule(l){2-9} 
                        & Accuracy & Rejections ($\downarrow$) & Accuracy               & Rejections              & Accuracy               & Rejections               & Accuracy              & Rejections             \\ \midrule
P0 (`Generated')        & 0.852    & 26                        & 0.769                  & 23              & 0.561                  & 6                & 0.592                 & 5              \\
P0 (`Fake')             & 0.849    & 45                        & 0.761                  & 49              & 0.543                  & 8                & 0.588                 & 5              \\ \bottomrule
\end{tabular}
\caption{Changing `fake' to `generated' in the prompt can lower rejection rates and improve accuracy for all models.
}
\label{tab:generated_or_fake}
\end{table*}

%% file: sections/5_limitations.tex
\section{Limitations}
\label{sec:limitations}

AI-generated content introduces significant ethical risks, particularly in misinformation, identity manipulation, and fraud. Robust detection systems are essential to address these concerns.
Ethically, ensuring transparency and accountability in detection models is critical, especially in sensitive areas like forensics and law enforcement. 
These models must be interpretable and free from biases that could lead to unfair outcomes. Privacy and security concerns also arise from the use of AI in surveillance and public discourse, requiring strict safeguards.

While MLLMs show promise in detecting AI-generated images, challenges remain in interpretability and alignment with human perception. Further research is needed to quantify this alignment and assess how human evaluators interpret MLLM-generated explanations in forensic applications.
Future research could explore adaptive fusion mechanisms that dynamically adjust reasoning strategies based on image characteristics, as well as adversarial training on high-quality AI-generated images to enhance model robustness.

%% file: sections/6_conclusion.tex
\section{Conclusion}
\label{sec:conclusion}

In this work, we explore the use of MLLMs for AI-generated image detection, offering a human-interpretable classification method. We introduce six detection paradigms tailored for MLLMs and validate their effectiveness across diverse image generator architectures. A majority vote from these paradigms can surpass traditional classification methods and the most accurate humans. The fusion of model responses, when passed again to the MLLM, allows for aggregated and accurate reasoning of why an image is considered real or AI-generated. Combining diverse heuristics and reasoning improves both accuracy and reliability. Our ablation study confirms that leveraging multiple detection paradigms outperforms single-method classification.

%% file: sections/7_appendix.tex
\section*{Appendix}

\input{figures/appendices/p0_open_source}

\section{Dataset Distribution}

\begin{figure}[h]
    \centering
    \includegraphics[width=\linewidth]{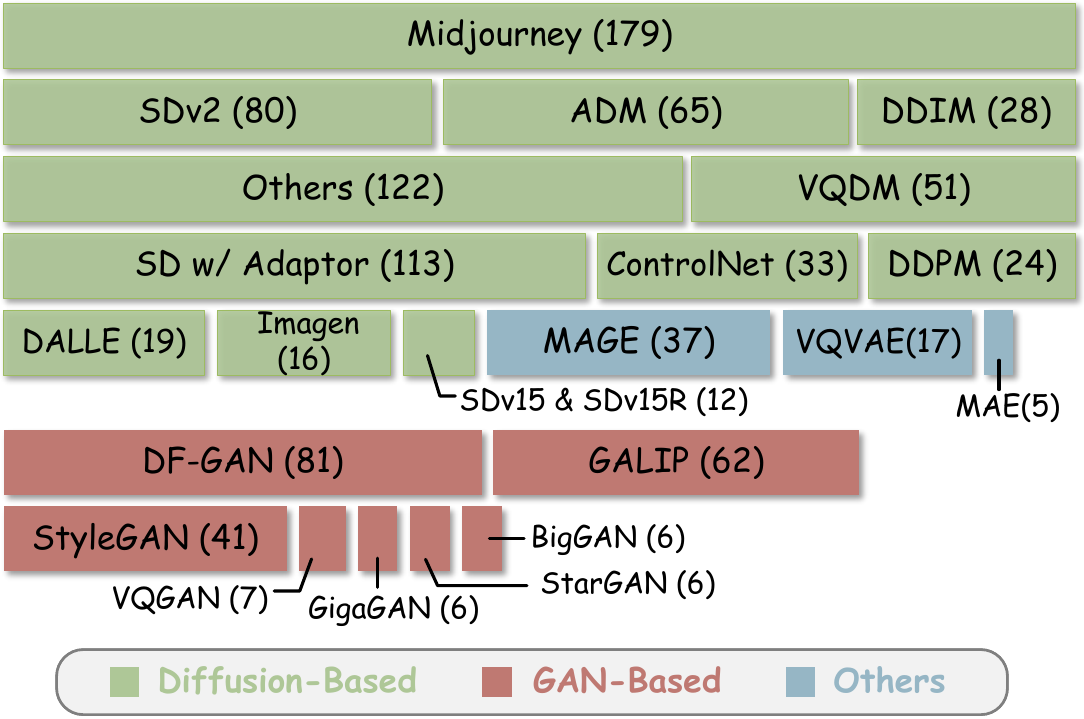}
    \Description{A bar chart showing the distribution of image generation methods. Most images are diffusion-based, followed by GAN-based and other architectures.}
    \caption{Distribution of AI generation methods among images in the dataset.}
    \label{fig:generators_dist}
\end{figure}

To ensure diversity, we curated a dataset containing 2000 images from multiple sources, covering three major architectures and variations of specific models, as detailed below.

\noindent \textbf{Diffusion-Based Generators.} We sampled from Midjourney  \citep{Midjourney}, SDv2  \citep{SD21}, ADM  \citep{ADM}, DDIM  \citep{DDIM}, DDPM \citep{DDPM}, Imagen \citep{Imagen}, VQDM  \citep{VQDM}, ControlNet  \citep{ControlNet}, SD with adaptors (LoRA  \citep{SDLora}, LyCORIS \citep{SDLyCORIS}) and DALLE 3 \citep{DALLE3}.

\noindent \textbf{GAN-Based Generators.} We sampled from DF-GAN \citep{DFGAN}, GALIP \citep{GALIP}, StyleGAN \citep{StyleGAN}, VQGAN \citep{VQGAN}, GigaGAN \citep{GigaGAN}, BigGAN \citep{BigGAN} and StarGAN \citep{StarGAN, StarGAN2}.

\noindent \textbf{Other Generators.} We sampled from VQVAE  \citep{VQVAE}, MAGE  \citep{MAGE} and MAE \citep{MAE}.

\noindent \textbf{Real Images.} We select real images from COCO \citep{COCO}, LAION-5B \citep{LAION5B}, 
LSUN Church \citep{LSUN}, FFHQ \citep{StyleGAN}, CelebA-HQ \citep{CelebAHQ}, ImageNet \citep{ImageNet} and AFHQ \citep{StarGAN2}.

\section{Discussions on Model Responses}

\begin{figure*}
    \centering
    \includegraphics[height=1\textheight]{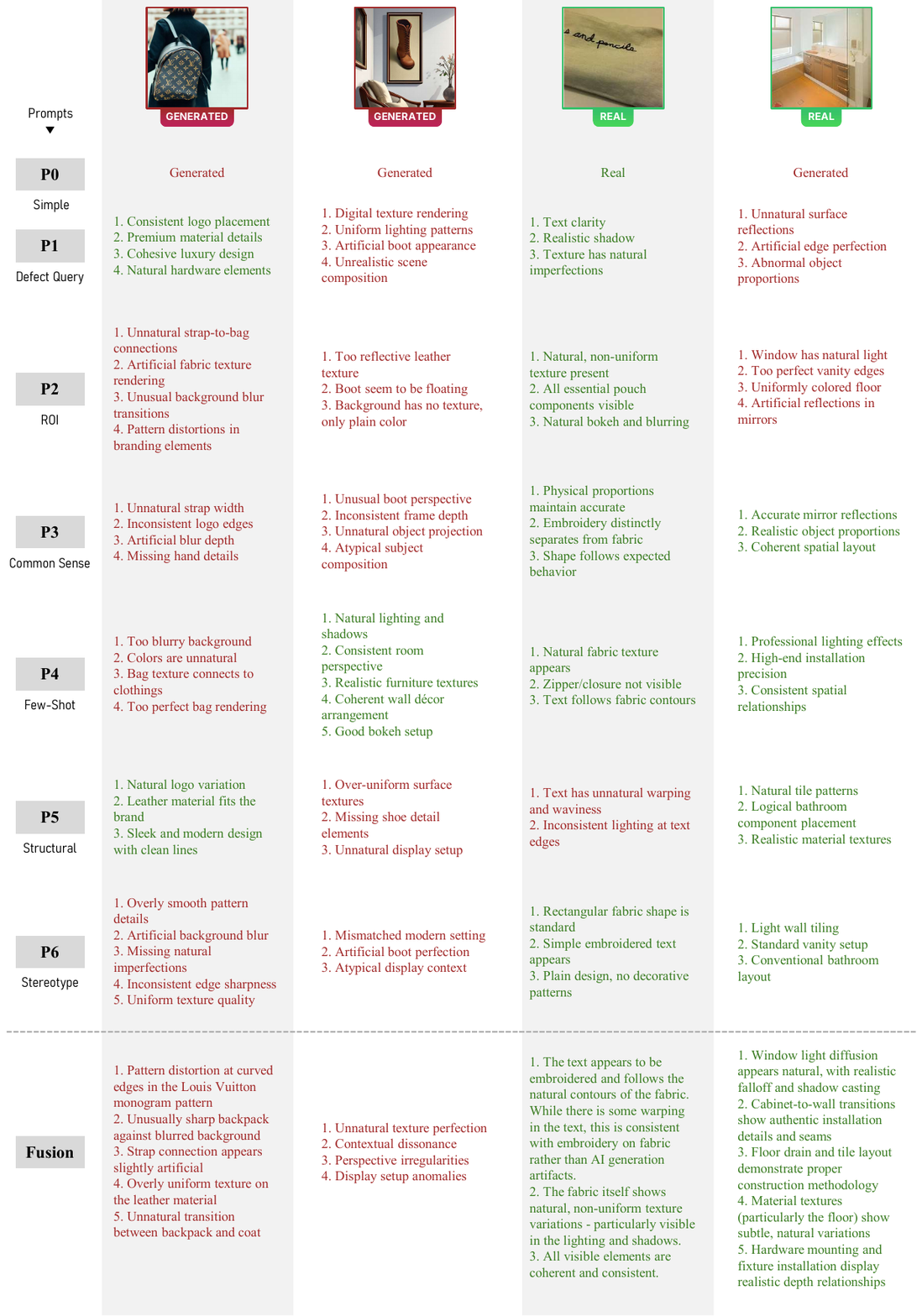}
    \Description{Post-fusion responses from GPT-4o for more images.}
    \caption{More examples with GPT-4o. Responses are summarized. Different colors of the text indicate the verdict for each query (red/green: fake/real).}
    \label{fig:more-examples}
\end{figure*}

To accommodate paper length constraints, we present additional results for qualitative analysis in~\cref{fig:more-examples}. These examples are unfiltered, showcasing the model’s typical performance.
Our analysis shows that GPT-4o produces varying verdicts across different prompts but maintains consistent reasoning accuracy during the fusion stage. This aligns with our experimental findings, where post-fusion responses achieve the highest accuracy among MLLM-based approaches. Our full prompts can be accessed from \href{https://anonymous.4open.science/r/mllm-defake/prompts}{this link}.

For conciseness, responses are further shortened with the prompt below.

\lstset{
    basicstyle=\ttfamily\footnotesize,
    breaklines=true,
    breakatwhitespace=true,
    tabsize=2,
    showspaces=false,
    showstringspaces=false
}
\begin{lstlisting}[label={lst:summarizing}]
<original_response>
---
Summarize the above response into no more than 5 key points, where each point should be fewer than 6 words. Output a numbered list in markdown format. Do not include the final verdict in your response.
\end{lstlisting}

\begin{figure}[ht]
    \centering
    \includegraphics[width=\linewidth]{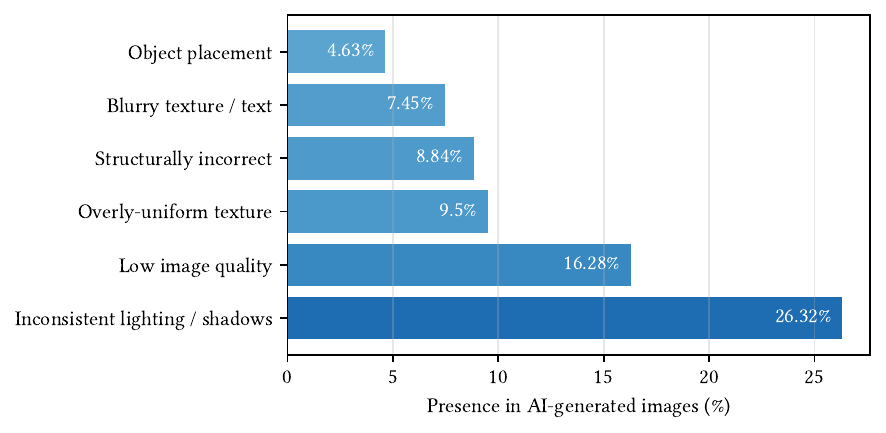}
    \Description{A bar chart listing features and their percentages found in AI-generated images.}
    \caption{Common key words present in model responses.}
    \label{fig:few-shot-variation}
\end{figure}

\textbf{Key words that model uses.} To qualitatively assess the model's attention tendency, we list the most common key words that are seen many times in model responses. While some of the model responses 

\section{Experimental Results on Other MLLMs}

We evaluate several MLLMs to distinguish AI-generated images from real ones. The results, as shown in \cref{tab:p0_open_source}, indicate that Qwen2-VL \citep{Qwen2VL} and QVQ \citep{QVQ} perform relatively well in this task. 
However, Qwen2-VL exhibits a bias towards classifying images as real, while QVQ, with 72B parameters, can take almost ten minutes for one fusion query on 4x NVIDIA H800 GPU.
Qwen2.5-VL-7B-Instruct also gives remarkable results, but it lacks the multi-turn image input chat template, rendering P2 and P4 unusable.
Gemini 1.5 Pro \citep{Gemini15}, as a proprietary model, scored lower than GPT-4o-mini in all three metrics.
We selected LLaVA-CoT \citep{LLaVACoT} for its reinforced CoT capabilities and conducted further experiments to explore the potential of open-source MLLMs in AI-generated image detection. Llama 3.2 Vision Instruct, the base model of LLaVA-CoT, is used for comparison.

\begin{figure}[htbp]
    \centering
    \includegraphics[width=\linewidth]{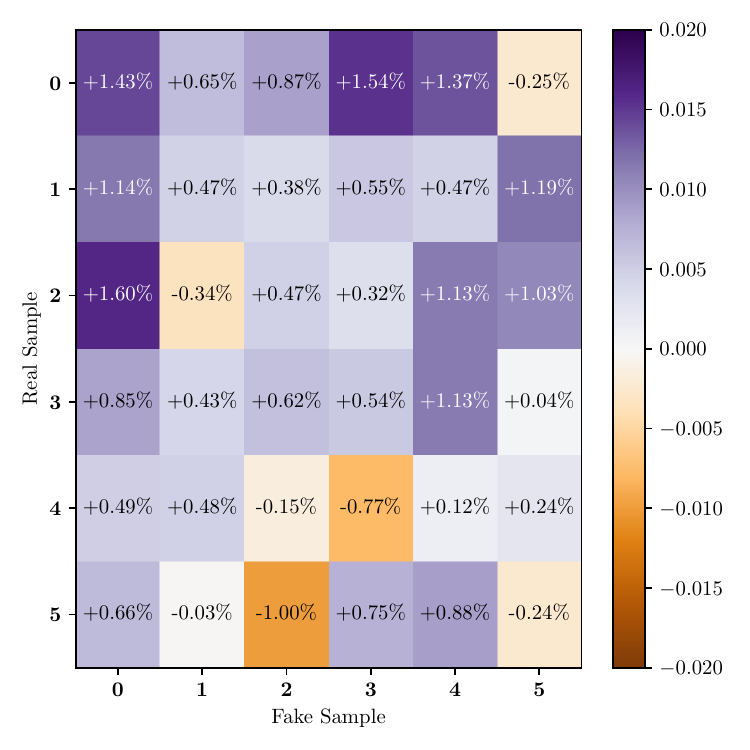}
    \Description{A grid showcasing the performance drift of few-shot prompt based on exemplar selection.}
    \caption{Grid representation of accuracy changes for different real-fake exemplar pairs in the few-shot learning setting. Positive values indicate accuracy improvements over the zero-shot baseline (P-Zero), while negative values denote declines.}
    \label{fig:few-shot-variation}
\end{figure}

\begin{figure*}[ht]
    \centering
    \includegraphics[width=\linewidth]{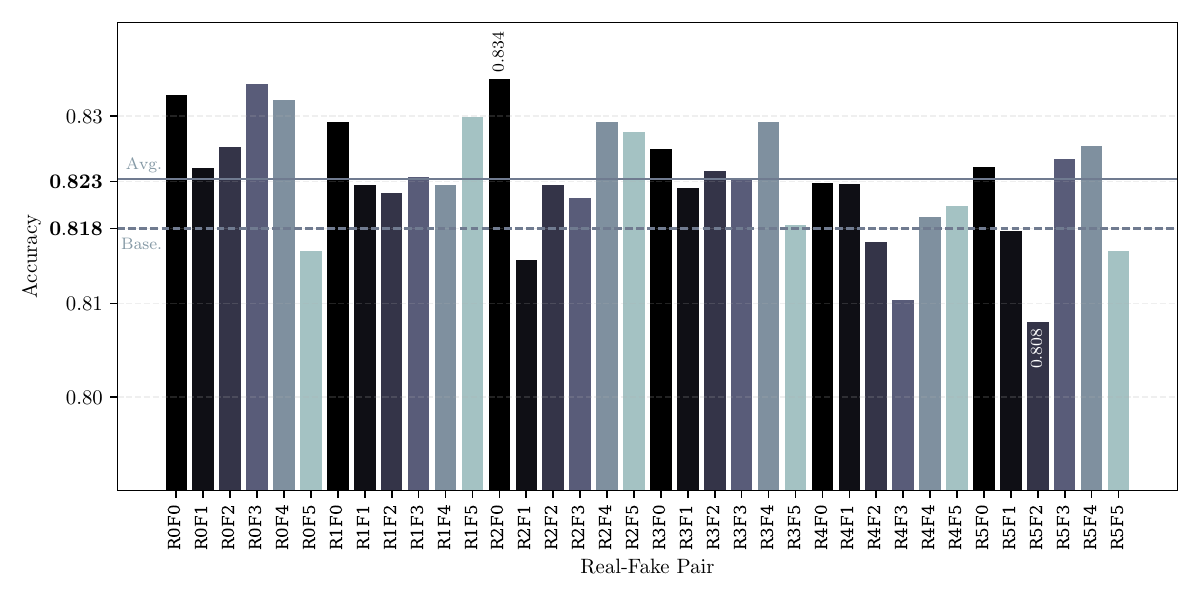}
    \Description{A bar chart showing the accuracy of few-shot prompt on different exemplar combinations.}
    \caption{Impact of real and fake sample selection on the overall accuracy of GPT-4o. 
    The \textit{dashed horizontal line} score represents zero-shot accuracy, while the \textit{solid horizontal line} line denotes the average accuracy across all two-shot pairs.}
    \label{fig:few-shot-bar}
\end{figure*}

\section{Discussions on Few-Shot Exemplar Selection}

MLLMs exhibit sensitivity to exemplar selection in few-shot learning \citep{VLFewShotExamples}. To systematically assess its impact, we compare few-shot learning (P3) against the zero-shot paradigm (P0), evaluating how exemplar choice affects model accuracy and decision stability.

We selected six real and six AI-generated images uniformly. These images were independently annotated by human researchers, identifying artifacts, inconsistencies, and perceptual anomalies that might influence MLLM classification. 
All 36 possible real-fake combinations were tested on the same dataset as used in Section~\ref{sec:experiments}. 
Additionally, we removed both exemplars from the prompt to create a zero-shot context (P-Zero), serving as the control group.

\cref{fig:few-shot-bar} presents the comparative accuracy of different few-shot sample combinations against the zero-shot baseline. Additionally, \cref{fig:few-shot-variation} provides a grid representation of accuracy variations across different pairings of real and fake exemplars.

From the results, we observe that model accuracy fluctuates depending on the selection of exemplars. The highest observed accuracy gain was +1.60\%, while the largest drop was -1.00\%. 
Despite these variations, the overall standard deviation remains moderate, indicating that while sample selection plays a role, GPT-4o exhibits relative stability in performance across different trials.

While sample selection influences performance, the variance remains limited, suggesting GPT-4o exhibits robustness to exemplar variations. In applications demanding high interpretability and reliability, exploring broader sample combinations could further optimize performance.

%% file: figures/appendices/p0_open_source.tex
\begin{table*}[htbp]
\centering
\resizebox{\textwidth}{!}{
\renewcommand{\arraystretch}{1.00}

\begin{tabular}{@{}ccccccccccccc@{}}
\toprule
\multirow{2}{*}{Prompt} &
  \multicolumn{3}{c}{\textbf{GPT-4o}} &
  \multicolumn{3}{c}{\textbf{GPT-4o-mini}} &
  \multicolumn{3}{c}{\textbf{Gemini 1.5 Pro}} &
  \multicolumn{3}{c}{\textbf{Kimi-VL-A3B-Instruct}} \\ \cmidrule(l){2-13} 
 &
  All &
  Real &
  \multicolumn{1}{c|}{Generated} &
  All &
  Real &
  \multicolumn{1}{c|}{Generated} &
  All &
  Real &
  \multicolumn{1}{c|}{Generated} &
  All &
  Real &
  Generated \\ \midrule
P0 &
  0.852 &
  0.855 &
  0.849 &
  0.769 &
  0.910 &
  0.628 &
  0.758 &
  0.906 &
  0.610 &
  0.585 &
  0.622 &
  0.548 \\ \midrule
\multirow{2}{*}{} &
  \multicolumn{3}{c}{\textbf{MiMo-VL-7B}} &
  \multicolumn{3}{c}{\textbf{Qwen2-VL-7B-Instruct}} &
  \multicolumn{3}{c}{\textbf{QVQ-72B-Preview}} &
  \multicolumn{3}{c}{\textbf{Qwen2.5-VL-7B-Instruct}} \\ \cmidrule(l){2-13} 
 &
  All &
  Real &
  \multicolumn{1}{c|}{Generated} &
  All &
  Real &
  \multicolumn{1}{c|}{Generated} &
  All &
  Real &
  \multicolumn{1}{c|}{Generated} &
  All &
  Real &
  Generated \\ \midrule
P0 &
  0.585 &
  0.887 &
  0.283 &
  0.623 &
  \textbf{0.901} &
  0.346 &
  \textbf{0.643} &
  0.789 &
  0.497 &
  0.621 &
  0.660 &
  0.582 \\ \midrule
\multirow{2}{*}{} &
  \multicolumn{3}{c}{\textbf{Ovis1.6-Gemma2-9B}} &
  \multicolumn{3}{c}{\textbf{InternVL2-8B-MPO}} &
  \multicolumn{3}{c}{\textbf{InternVL2.5-26B}} &
  \multicolumn{3}{c}{\textbf{MiniCPM-V-2.6 (8B)}} \\ \cmidrule(l){2-13} 
 &
  All &
  Real &
  \multicolumn{1}{c|}{Generated} &
  All &
  Real &
  \multicolumn{1}{c|}{Generated} &
  All &
  Real &
  \multicolumn{1}{c|}{Generated} &
  All &
  Real &
  Generated \\ \midrule
P0 &
  0.559 &
  0.828 &
  0.290 &
  0.592 &
  0.623 &
  \textbf{0.561} &
  0.566 &
  0.673 &
  0.460 &
  0.545 &
  0.766 &
  0.324 \\ \bottomrule
\end{tabular}%

}
\caption{Experimental result of P0 on different open-source and proprietary models.}
\label{tab:p0_open_source}
\end{table*}

%% file: main.bbl

\begin{thebibliography}{84}


\ifx \showCODEN    \undefined \def \showCODEN     #1{\unskip}     \fi
\ifx \showISBNx    \undefined \def \showISBNx     #1{\unskip}     \fi
\ifx \showISBNxiii \undefined \def \showISBNxiii  #1{\unskip}     \fi
\ifx \showISSN     \undefined \def \showISSN      #1{\unskip}     \fi
\ifx \showLCCN     \undefined \def \showLCCN      #1{\unskip}     \fi
\ifx \shownote     \undefined \def \shownote      #1{#1}          \fi
\ifx \showarticletitle \undefined \def \showarticletitle #1{#1}   \fi
\ifx \showURL      \undefined \def \showURL       {\relax}        \fi
\providecommand\bibfield[2]{#2}
\providecommand\bibinfo[2]{#2}
\providecommand\natexlab[1]{#1}
\providecommand\showeprint[2][]{arXiv:#2}

\bibitem[Bai et~al\mbox{.}(2023)]%
        {QwenVL}
\bibfield{author}{\bibinfo{person}{Jinze Bai}, \bibinfo{person}{Shuai Bai},
  \bibinfo{person}{Shusheng Yang}, \bibinfo{person}{Shijie Wang},
  \bibinfo{person}{Sinan Tan}, \bibinfo{person}{Peng Wang},
  \bibinfo{person}{Junyang Lin}, \bibinfo{person}{Chang Zhou}, {and}
  \bibinfo{person}{Jingren Zhou}.} \bibinfo{year}{2023}\natexlab{}.
\newblock \showarticletitle{Qwen-VL: A Versatile Vision-Language Model for
  Understanding, Localization, Text Reading, and Beyond}.
\newblock \bibinfo{journal}{\emph{arXiv preprint arXiv:2308.12966}}
  (\bibinfo{year}{2023}).
\newblock


\bibitem[Bavishi et~al\mbox{.}(2023)]%
        {Fuyu8B}
\bibfield{author}{\bibinfo{person}{Rohan Bavishi}, \bibinfo{person}{Erich
  Elsen}, \bibinfo{person}{Curtis Hawthorne}, \bibinfo{person}{Maxwell Nye},
  \bibinfo{person}{Augustus Odena}, \bibinfo{person}{Arushi Somani}, {and}
  \bibinfo{person}{Sa\u{g}nak Ta\c{s}\i{}rlar}.}
  \bibinfo{year}{2023}\natexlab{}.
\newblock \bibinfo{title}{Introducing our Multimodal Models}.
\newblock
\urldef\tempurl%
\url{https://www.adept.ai/blog/fuyu-8b}
\showURL{%
\tempurl}


\bibitem[Bi et~al\mbox{.}(2023)]%
        {DefakeByReals}
\bibfield{author}{\bibinfo{person}{Xiuli Bi}, \bibinfo{person}{Bo Liu},
  \bibinfo{person}{Fan Yang}, \bibinfo{person}{Bin Xiao},
  \bibinfo{person}{Weisheng Li}, \bibinfo{person}{Gao Huang}, {and}
  \bibinfo{person}{Pamela~C. Cosman}.} \bibinfo{year}{2023}\natexlab{}.
\newblock \bibinfo{title}{Detecting Generated Images by Real Images Only}.
\newblock
\showeprint[arxiv]{2311.00962}~[cs.CV]
\urldef\tempurl%
\url{https://arxiv.org/abs/2311.00962}
\showURL{%
\tempurl}


\bibitem[Brock et~al\mbox{.}(2018)]%
        {BigGAN}
\bibfield{author}{\bibinfo{person}{Andrew Brock}, \bibinfo{person}{Jeff
  Donahue}, {and} \bibinfo{person}{Karen Simonyan}.}
  \bibinfo{year}{2018}\natexlab{}.
\newblock \showarticletitle{Large Scale GAN Training for High Fidelity Natural
  Image Synthesis}. In \bibinfo{booktitle}{\emph{International Conference on
  Learning Representations}}.
\newblock


\bibitem[Chai et~al\mbox{.}(2020)]%
        {ArtifactWhatsDetectable}
\bibfield{author}{\bibinfo{person}{Lucy Chai}, \bibinfo{person}{David Bau},
  \bibinfo{person}{Ser-Nam Lim}, {and} \bibinfo{person}{Phillip Isola}.}
  \bibinfo{year}{2020}\natexlab{}.
\newblock \showarticletitle{What Makes Fake Images Detectable? Understanding
  Properties that Generalize}. In \bibinfo{booktitle}{\emph{Computer Vision –
  ECCV 2020: 16th European Conference, Glasgow, UK, August 23–28, 2020,
  Proceedings, Part XXVI}} (Glasgow, United Kingdom).
  \bibinfo{publisher}{Springer-Verlag}, \bibinfo{address}{Berlin, Heidelberg},
  \bibinfo{pages}{103–120}.
\newblock
\showISBNx{978-3-030-58573-0}
\href{https://doi.org/10.1007/978-3-030-58574-7_7}{doi:\nolinkurl{10.1007/978-3-030-58574-7_7}}


\bibitem[Chen et~al\mbox{.}(2015)]%
        {COCO}
\bibfield{author}{\bibinfo{person}{Xinlei Chen}, \bibinfo{person}{Hao Fang},
  \bibinfo{person}{Tsung-Yi Lin}, \bibinfo{person}{Ramakrishna Vedantam},
  \bibinfo{person}{Saurabh Gupta}, \bibinfo{person}{Piotr Doll{\'a}r}, {and}
  \bibinfo{person}{C.~Lawrence Zitnick}.} \bibinfo{year}{2015}\natexlab{}.
\newblock \showarticletitle{Microsoft COCO Captions: Data Collection and
  Evaluation Server}.
\newblock \bibinfo{journal}{\emph{ArXiv}}  \bibinfo{volume}{abs/1504.00325}
  (\bibinfo{year}{2015}).
\newblock
\urldef\tempurl%
\url{https://api.semanticscholar.org/CorpusID:2210455}
\showURL{%
\tempurl}


\bibitem[Chen et~al\mbox{.}(2024)]%
        {InternVL}
\bibfield{author}{\bibinfo{person}{Zhe Chen}, \bibinfo{person}{Jiannan Wu},
  \bibinfo{person}{Wenhai Wang}, \bibinfo{person}{Weijie Su},
  \bibinfo{person}{Guo Chen}, \bibinfo{person}{Sen Xing},
  \bibinfo{person}{Muyan Zhong}, \bibinfo{person}{Qinglong Zhang},
  \bibinfo{person}{Xizhou Zhu}, \bibinfo{person}{Lewei Lu}, {et~al\mbox{.}}}
  \bibinfo{year}{2024}\natexlab{}.
\newblock \showarticletitle{Internvl: Scaling up vision foundation models and
  aligning for generic visual-linguistic tasks}. In
  \bibinfo{booktitle}{\emph{Proceedings of the IEEE/CVF Conference on Computer
  Vision and Pattern Recognition}}. \bibinfo{pages}{24185--24198}.
\newblock


\bibitem[Choi et~al\mbox{.}(2018)]%
        {StarGAN}
\bibfield{author}{\bibinfo{person}{Yunjey Choi}, \bibinfo{person}{Minje Choi},
  \bibinfo{person}{Munyoung Kim}, \bibinfo{person}{Jung-Woo Ha},
  \bibinfo{person}{Sunghun Kim}, {and} \bibinfo{person}{Jaegul Choo}.}
  \bibinfo{year}{2018}\natexlab{}.
\newblock \showarticletitle{Stargan: Unified generative adversarial networks
  for multi-domain image-to-image translation}. In
  \bibinfo{booktitle}{\emph{Proceedings of the IEEE conference on computer
  vision and pattern recognition}}. \bibinfo{pages}{8789--8797}.
\newblock


\bibitem[Choi et~al\mbox{.}(2020)]%
        {StarGAN2}
\bibfield{author}{\bibinfo{person}{Yunjey Choi}, \bibinfo{person}{Youngjung
  Uh}, \bibinfo{person}{Jaejun Yoo}, {and} \bibinfo{person}{Jung-Woo Ha}.}
  \bibinfo{year}{2020}\natexlab{}.
\newblock \showarticletitle{Stargan v2: Diverse image synthesis for multiple
  domains}. In \bibinfo{booktitle}{\emph{Proceedings of the IEEE/CVF conference
  on computer vision and pattern recognition}}. \bibinfo{pages}{8188--8197}.
\newblock


\bibitem[Corvi et~al\mbox{.}(2023)]%
        {DMImageDetection}
\bibfield{author}{\bibinfo{person}{Riccardo Corvi}, \bibinfo{person}{Davide
  Cozzolino}, \bibinfo{person}{Giada Zingarini}, \bibinfo{person}{Giovanni
  Poggi}, \bibinfo{person}{Koki Nagano}, {and} \bibinfo{person}{Luisa
  Verdoliva}.} \bibinfo{year}{2023}\natexlab{}.
\newblock \showarticletitle{On The Detection of Synthetic Images Generated by
  Diffusion Models}. In \bibinfo{booktitle}{\emph{IEEE International Conference
  on Acoustics, Speech and Signal Processing (ICASSP)}}. \bibinfo{pages}{1--5}.
\newblock
\href{https://doi.org/10.1109/ICASSP49357.2023.10095167}{doi:\nolinkurl{10.1109/ICASSP49357.2023.10095167}}


\bibitem[Deng et~al\mbox{.}(2009)]%
        {ImageNet}
\bibfield{author}{\bibinfo{person}{Jia Deng}, \bibinfo{person}{Wei Dong},
  \bibinfo{person}{Richard Socher}, \bibinfo{person}{Li-Jia Li},
  \bibinfo{person}{K. Li}, {and} \bibinfo{person}{Li Fei-Fei}.}
  \bibinfo{year}{2009}\natexlab{}.
\newblock \showarticletitle{ImageNet: A large-scale hierarchical image
  database}.
\newblock \bibinfo{journal}{\emph{2009 IEEE Conference on Computer Vision and
  Pattern Recognition}} (\bibinfo{year}{2009}), \bibinfo{pages}{248--255}.
\newblock
\urldef\tempurl%
\url{https://api.semanticscholar.org/CorpusID:57246310}
\showURL{%
\tempurl}


\bibitem[Dhariwal and Nichol(2021a)]%
        {Diffusion}
\bibfield{author}{\bibinfo{person}{Prafulla Dhariwal} {and}
  \bibinfo{person}{Alex Nichol}.} \bibinfo{year}{2021}\natexlab{a}.
\newblock \showarticletitle{Diffusion Models Beat GANs on Image Synthesis}.
\newblock \bibinfo{journal}{\emph{ArXiv}}  \bibinfo{volume}{abs/2105.05233}
  (\bibinfo{year}{2021}).
\newblock
\urldef\tempurl%
\url{https://api.semanticscholar.org/CorpusID:234357997}
\showURL{%
\tempurl}


\bibitem[Dhariwal and Nichol(2021b)]%
        {ADM}
\bibfield{author}{\bibinfo{person}{Prafulla Dhariwal} {and}
  \bibinfo{person}{Alexander Nichol}.} \bibinfo{year}{2021}\natexlab{b}.
\newblock \showarticletitle{Diffusion models beat gans on image synthesis}.
\newblock \bibinfo{journal}{\emph{Advances in neural information processing
  systems}}  \bibinfo{volume}{34} (\bibinfo{year}{2021}),
  \bibinfo{pages}{8780--8794}.
\newblock


\bibitem[Dzanic et~al\mbox{.}(2020)]%
        {ArtifactSpectrumDefake}
\bibfield{author}{\bibinfo{person}{Tarik Dzanic}, \bibinfo{person}{Karan Shah},
  {and} \bibinfo{person}{Freddie~D. Witherden}.}
  \bibinfo{year}{2020}\natexlab{}.
\newblock \showarticletitle{Fourier spectrum discrepancies in deep network
  generated images}. In \bibinfo{booktitle}{\emph{Proceedings of the 34th
  International Conference on Neural Information Processing Systems}}
  (Vancouver, BC, Canada) \emph{(\bibinfo{series}{NIPS '20})}.
  \bibinfo{publisher}{Curran Associates Inc.}, \bibinfo{address}{Red Hook, NY,
  USA}, Article \bibinfo{articleno}{254}, \bibinfo{numpages}{11}~pages.
\newblock
\showISBNx{9781713829546}


\bibitem[Esser et~al\mbox{.}(2024)]%
        {SD3}
\bibfield{author}{\bibinfo{person}{Patrick Esser}, \bibinfo{person}{Sumith
  Kulal}, \bibinfo{person}{Andreas Blattmann}, \bibinfo{person}{Rahim
  Entezari}, \bibinfo{person}{Jonas Müller}, \bibinfo{person}{Harry Saini},
  \bibinfo{person}{Yam Levi}, \bibinfo{person}{Dominik Lorenz},
  \bibinfo{person}{Axel Sauer}, \bibinfo{person}{Frederic Boesel},
  \bibinfo{person}{Dustin Podell}, \bibinfo{person}{Tim Dockhorn},
  \bibinfo{person}{Zion English}, \bibinfo{person}{Kyle Lacey},
  \bibinfo{person}{Alex Goodwin}, \bibinfo{person}{Yannik Marek}, {and}
  \bibinfo{person}{Robin Rombach}.} \bibinfo{year}{2024}\natexlab{}.
\newblock \bibinfo{title}{Scaling Rectified Flow Transformers for
  High-Resolution Image Synthesis}.
\newblock
\showeprint[arxiv]{2403.03206}~[cs.CV]
\urldef\tempurl%
\url{https://arxiv.org/abs/2403.03206}
\showURL{%
\tempurl}


\bibitem[Esser et~al\mbox{.}(2021)]%
        {VQGAN}
\bibfield{author}{\bibinfo{person}{Patrick Esser}, \bibinfo{person}{Robin
  Rombach}, {and} \bibinfo{person}{Bjorn Ommer}.}
  \bibinfo{year}{2021}\natexlab{}.
\newblock \showarticletitle{Taming transformers for high-resolution image
  synthesis}. In \bibinfo{booktitle}{\emph{Proceedings of the IEEE/CVF
  conference on computer vision and pattern recognition}}.
  \bibinfo{pages}{12873--12883}.
\newblock


\bibitem[Frank et~al\mbox{.}(2020)]%
        {ArtifactUseFreq}
\bibfield{author}{\bibinfo{person}{Joel Frank}, \bibinfo{person}{Thorsten
  Eisenhofer}, \bibinfo{person}{Lea Sch\"{o}nherr}, \bibinfo{person}{Asja
  Fischer}, \bibinfo{person}{Dorothea Kolossa}, {and} \bibinfo{person}{Thorsten
  Holz}.} \bibinfo{year}{2020}\natexlab{}.
\newblock \showarticletitle{Leveraging frequency analysis for deep fake image
  recognition}. In \bibinfo{booktitle}{\emph{Proceedings of the 37th
  International Conference on Machine Learning}}
  \emph{(\bibinfo{series}{ICML'20})}. \bibinfo{publisher}{JMLR.org}, Article
  \bibinfo{articleno}{304}, \bibinfo{numpages}{12}~pages.
\newblock


\bibitem[Gao et~al\mbox{.}(2025)]%
        {FakeReasoning}
\bibfield{author}{\bibinfo{person}{Yueying Gao}, \bibinfo{person}{Dongliang
  Chang}, \bibinfo{person}{Bingyao Yu}, \bibinfo{person}{Haotian Qin},
  \bibinfo{person}{Lei Chen}, \bibinfo{person}{Kongming Liang}, {and}
  \bibinfo{person}{Zhanyu Ma}.} \bibinfo{year}{2025}\natexlab{}.
\newblock \showarticletitle{FakeReasoning: Towards Generalizable Forgery
  Detection and Reasoning}.
\newblock \bibinfo{journal}{\emph{arXiv preprint arXiv:2503.21210}}
  (\bibinfo{year}{2025}).
\newblock
\urldef\tempurl%
\url{https://arxiv.org/abs/2503.21210}
\showURL{%
\tempurl}


\bibitem[Gao et~al\mbox{.}(2024)]%
        {HallucinationLLM}
\bibfield{author}{\bibinfo{person}{Yifei Gao}, \bibinfo{person}{Jiaqi Wang},
  \bibinfo{person}{Zhiyu Lin}, {and} \bibinfo{person}{Jitao Sang}.}
  \bibinfo{year}{2024}\natexlab{}.
\newblock \bibinfo{title}{AIGCs Confuse AI Too: Investigating and Explaining
  Synthetic Image-induced Hallucinations in Large Vision-Language Models}.
\newblock
\showeprint[arxiv]{2403.08542}~[cs.CV]
\urldef\tempurl%
\url{https://arxiv.org/abs/2403.08542}
\showURL{%
\tempurl}


\bibitem[{Gemini Team}(2024a)]%
        {Gemini15}
\bibfield{author}{\bibinfo{person}{{Gemini Team}}.}
  \bibinfo{year}{2024}\natexlab{a}.
\newblock \bibinfo{title}{Gemini 1.5: Unlocking multimodal understanding across
  millions of tokens of context}.
\newblock
\showeprint[arxiv]{2403.05530}~[cs.CL]
\urldef\tempurl%
\url{https://arxiv.org/abs/2403.05530}
\showURL{%
\tempurl}


\bibitem[{Gemini Team}(2024b)]%
        {Gemini}
\bibfield{author}{\bibinfo{person}{{Gemini Team}}.}
  \bibinfo{year}{2024}\natexlab{b}.
\newblock \bibinfo{title}{Gemini: A Family of Highly Capable Multimodal
  Models}.
\newblock
\showeprint[arxiv]{2312.11805}~[cs.CL]
\urldef\tempurl%
\url{https://arxiv.org/abs/2312.11805}
\showURL{%
\tempurl}


\bibitem[Goodfellow et~al\mbox{.}(2014)]%
        {GAN}
\bibfield{author}{\bibinfo{person}{Ian Goodfellow}, \bibinfo{person}{Jean
  Pouget-Abadie}, \bibinfo{person}{Mehdi Mirza}, \bibinfo{person}{Bing Xu},
  \bibinfo{person}{David Warde-Farley}, \bibinfo{person}{Sherjil Ozair},
  \bibinfo{person}{Aaron Courville}, {and} \bibinfo{person}{Y. Bengio}.}
  \bibinfo{year}{2014}\natexlab{}.
\newblock \showarticletitle{Generative Adversarial Networks}.
\newblock \bibinfo{journal}{\emph{Advances in Neural Information Processing
  Systems}}  \bibinfo{volume}{3} (\bibinfo{date}{06} \bibinfo{year}{2014}).
\newblock
\href{https://doi.org/10.1145/3422622}{doi:\nolinkurl{10.1145/3422622}}


\bibitem[Grommelt et~al\mbox{.}(2024)]%
        {FakeOrJPEG}
\bibfield{author}{\bibinfo{person}{Patrick Grommelt}, \bibinfo{person}{Louis
  Weiss}, \bibinfo{person}{Franz-Josef Pfreundt}, {and} \bibinfo{person}{Janis
  Keuper}.} \bibinfo{year}{2024}\natexlab{}.
\newblock \showarticletitle{Fake or JPEG? Revealing Common Biases in Generated
  Image Detection Datasets}.
\newblock \bibinfo{journal}{\emph{ArXiv}}  \bibinfo{volume}{abs/2403.17608}
  (\bibinfo{year}{2024}).
\newblock
\urldef\tempurl%
\url{https://api.semanticscholar.org/CorpusID:268691905}
\showURL{%
\tempurl}


\bibitem[Gu et~al\mbox{.}(2022)]%
        {VQDM}
\bibfield{author}{\bibinfo{person}{Shuyang Gu}, \bibinfo{person}{Dong Chen},
  \bibinfo{person}{Jianmin Bao}, \bibinfo{person}{Fang Wen},
  \bibinfo{person}{Bo Zhang}, \bibinfo{person}{Dongdong Chen},
  \bibinfo{person}{Lu Yuan}, {and} \bibinfo{person}{Baining Guo}.}
  \bibinfo{year}{2022}\natexlab{}.
\newblock \showarticletitle{Vector quantized diffusion model for text-to-image
  synthesis}. In \bibinfo{booktitle}{\emph{Proceedings of the IEEE/CVF
  Conference on Computer Vision and Pattern Recognition}}.
  \bibinfo{pages}{10696--10706}.
\newblock


\bibitem[Guo et~al\mbox{.}(2024)]%
        {VLFewShotExamples}
\bibfield{author}{\bibinfo{person}{Zhaojun Guo}, \bibinfo{person}{Jinghui Lu},
  \bibinfo{person}{Xuejing Liu}, \bibinfo{person}{Rui Zhao},
  \bibinfo{person}{ZhenXing Qian}, {and} \bibinfo{person}{Fei Tan}.}
  \bibinfo{year}{2024}\natexlab{}.
\newblock \bibinfo{title}{What Makes Good Few-shot Examples for Vision-Language
  Models?}
\newblock
\showeprint[arxiv]{2405.13532}~[cs.CV]
\urldef\tempurl%
\url{https://arxiv.org/abs/2405.13532}
\showURL{%
\tempurl}


\bibitem[He et~al\mbox{.}(2022)]%
        {MAE}
\bibfield{author}{\bibinfo{person}{Kaiming He}, \bibinfo{person}{Xinlei Chen},
  \bibinfo{person}{Saining Xie}, \bibinfo{person}{Yanghao Li},
  \bibinfo{person}{Piotr Doll{\'a}r}, {and} \bibinfo{person}{Ross Girshick}.}
  \bibinfo{year}{2022}\natexlab{}.
\newblock \showarticletitle{Masked autoencoders are scalable vision learners}.
  In \bibinfo{booktitle}{\emph{Proceedings of the IEEE/CVF conference on
  computer vision and pattern recognition}}. \bibinfo{pages}{16000--16009}.
\newblock


\bibitem[Ho et~al\mbox{.}(2020)]%
        {DDPM}
\bibfield{author}{\bibinfo{person}{Jonathan Ho}, \bibinfo{person}{Ajay Jain},
  {and} \bibinfo{person}{P. Abbeel}.} \bibinfo{year}{2020}\natexlab{}.
\newblock \showarticletitle{Denoising Diffusion Probabilistic Models}.
\newblock \bibinfo{journal}{\emph{ArXiv}}  \bibinfo{volume}{abs/2006.11239}
  (\bibinfo{year}{2020}).
\newblock
\urldef\tempurl%
\url{https://api.semanticscholar.org/CorpusID:219955663}
\showURL{%
\tempurl}


\bibitem[Holub(2022)]%
        {Midjourney}
\bibfield{author}{\bibinfo{person}{Oleksii Holub}.}
  \bibinfo{year}{2022}\natexlab{}.
\newblock \bibinfo{booktitle}{\emph{Midjourney}}.
\newblock
\urldef\tempurl%
\url{https://www.midjourney.com/home/}
\showURL{%
\tempurl}


\bibitem[Hong et~al\mbox{.}(2025)]%
        {WildFake}
\bibfield{author}{\bibinfo{person}{Yan Hong}, \bibinfo{person}{Jianming Feng},
  \bibinfo{person}{Haoxing Chen}, \bibinfo{person}{Jun Lan},
  \bibinfo{person}{Huijia Zhu}, \bibinfo{person}{Weiqiang Wang}, {and}
  \bibinfo{person}{Jianfu Zhang}.} \bibinfo{year}{2025}\natexlab{}.
\newblock \showarticletitle{Wildfake: A large-scale and hierarchical dataset
  for ai-generated images detection}. In \bibinfo{booktitle}{\emph{Proceedings
  of the AAAI Conference on Artificial Intelligence}},
  Vol.~\bibinfo{volume}{39}. \bibinfo{pages}{3500--3508}.
\newblock


\bibitem[Hu et~al\mbox{.}(2021)]%
        {SDLora}
\bibfield{author}{\bibinfo{person}{Edward~J Hu}, \bibinfo{person}{Yelong Shen},
  \bibinfo{person}{Phillip Wallis}, \bibinfo{person}{Zeyuan Allen-Zhu},
  \bibinfo{person}{Yuanzhi Li}, \bibinfo{person}{Shean Wang},
  \bibinfo{person}{Lu Wang}, {and} \bibinfo{person}{Weizhu Chen}.}
  \bibinfo{year}{2021}\natexlab{}.
\newblock \showarticletitle{Lora: Low-rank adaptation of large language
  models}.
\newblock \bibinfo{journal}{\emph{arXiv preprint arXiv:2106.09685}}
  (\bibinfo{year}{2021}).
\newblock


\bibitem[Kamoi et~al\mbox{.}(2024)]%
        {VisOnlyQA}
\bibfield{author}{\bibinfo{person}{Ryo Kamoi}, \bibinfo{person}{Yusen Zhang},
  \bibinfo{person}{Sarkar Snigdha~Sarathi Das}, \bibinfo{person}{Ranran~Haoran
  Zhang}, {and} \bibinfo{person}{Rui Zhang}.} \bibinfo{year}{2024}\natexlab{}.
\newblock \bibinfo{title}{VisOnlyQA: Large Vision Language Models Still
  Struggle with Visual Perception of Geometric Information}.
\newblock


\bibitem[Kang et~al\mbox{.}(2023)]%
        {GigaGAN}
\bibfield{author}{\bibinfo{person}{Minguk Kang}, \bibinfo{person}{Jun-Yan Zhu},
  \bibinfo{person}{Richard Zhang}, \bibinfo{person}{Jaesik Park},
  \bibinfo{person}{Eli Shechtman}, \bibinfo{person}{Sylvain Paris}, {and}
  \bibinfo{person}{Taesung Park}.} \bibinfo{year}{2023}\natexlab{}.
\newblock \showarticletitle{Scaling up gans for text-to-image synthesis}. In
  \bibinfo{booktitle}{\emph{Proceedings of the IEEE/CVF Conference on Computer
  Vision and Pattern Recognition}}. \bibinfo{pages}{10124--10134}.
\newblock


\bibitem[Karras et~al\mbox{.}(2017)]%
        {CelebAHQ}
\bibfield{author}{\bibinfo{person}{Tero Karras}, \bibinfo{person}{Timo Aila},
  \bibinfo{person}{Samuli Laine}, {and} \bibinfo{person}{Jaakko Lehtinen}.}
  \bibinfo{year}{2017}\natexlab{}.
\newblock \showarticletitle{Progressive growing of gans for improved quality,
  stability, and variation}.
\newblock \bibinfo{journal}{\emph{arXiv preprint arXiv:1710.10196}}
  (\bibinfo{year}{2017}).
\newblock


\bibitem[Karras et~al\mbox{.}(2018)]%
        {StyleGAN}
\bibfield{author}{\bibinfo{person}{Tero Karras}, \bibinfo{person}{Samuli
  Laine}, {and} \bibinfo{person}{Timo Aila}.} \bibinfo{year}{2018}\natexlab{}.
\newblock \showarticletitle{A Style-Based Generator Architecture for Generative
  Adversarial Networks}.
\newblock \bibinfo{journal}{\emph{2019 IEEE/CVF Conference on Computer Vision
  and Pattern Recognition (CVPR)}} (\bibinfo{year}{2018}),
  \bibinfo{pages}{4396--4405}.
\newblock
\urldef\tempurl%
\url{https://api.semanticscholar.org/CorpusID:54482423}
\showURL{%
\tempurl}


\bibitem[Katirai et~al\mbox{.}(2023)]%
        {SocialIssues}
\bibfield{author}{\bibinfo{person}{Amelia Katirai}, \bibinfo{person}{Noa
  Garc{\'i}a}, \bibinfo{person}{Kazuki Ide}, \bibinfo{person}{Yuta Nakashima},
  {and} \bibinfo{person}{Atsuo Kishimoto}.} \bibinfo{year}{2023}\natexlab{}.
\newblock \showarticletitle{Situating the social issues of image generation
  models in the model life cycle: a sociotechnical approach}.
\newblock \bibinfo{journal}{\emph{ArXiv}}  \bibinfo{volume}{abs/2311.18345}
  (\bibinfo{year}{2023}).
\newblock
\urldef\tempurl%
\url{https://api.semanticscholar.org/CorpusID:265506098}
\showURL{%
\tempurl}


\bibitem[Kwon et~al\mbox{.}(2023)]%
        {vllm}
\bibfield{author}{\bibinfo{person}{Woosuk Kwon}, \bibinfo{person}{Zhuohan Li},
  \bibinfo{person}{Siyuan Zhuang}, \bibinfo{person}{Ying Sheng},
  \bibinfo{person}{Lianmin Zheng}, \bibinfo{person}{Cody~Hao Yu},
  \bibinfo{person}{Joseph~E. Gonzalez}, \bibinfo{person}{Hao Zhang}, {and}
  \bibinfo{person}{Ion Stoica}.} \bibinfo{year}{2023}\natexlab{}.
\newblock \showarticletitle{Efficient Memory Management for Large Language
  Model Serving with PagedAttention}. In \bibinfo{booktitle}{\emph{Proceedings
  of the ACM SIGOPS 29th Symposium on Operating Systems Principles}}.
\newblock


\bibitem[Li et~al\mbox{.}(2023)]%
        {MAGE}
\bibfield{author}{\bibinfo{person}{Tianhong Li}, \bibinfo{person}{Huiwen
  Chang}, \bibinfo{person}{Shlok~Kumar Mishra}, \bibinfo{person}{Han Zhang},
  \bibinfo{person}{Dina Katabi}, {and} \bibinfo{person}{Dilip Krishnan}.}
  \bibinfo{year}{2023}\natexlab{}.
\newblock \showarticletitle{MAGE: MAsked Generative Encoder to Unify
  Representation Learning and Image Synthesis}. In
  \bibinfo{booktitle}{\emph{2023 IEEE/CVF Conference on Computer Vision and
  Pattern Recognition (CVPR)}}. \bibinfo{pages}{2142--2152}.
\newblock
\href{https://doi.org/10.1109/CVPR52729.2023.00213}{doi:\nolinkurl{10.1109/CVPR52729.2023.00213}}


\bibitem[Li et~al\mbox{.}(2024)]%
        {FakeBench}
\bibfield{author}{\bibinfo{person}{Yixuan Li}, \bibinfo{person}{Xuelin Liu},
  \bibinfo{person}{Xiaoyang Wang}, \bibinfo{person}{Shiqi Wang}, {and}
  \bibinfo{person}{Weisi Lin}.} \bibinfo{year}{2024}\natexlab{}.
\newblock \showarticletitle{FakeBench: Uncover the Achilles' Heels of Fake
  Images with Large Multimodal Models}.
\newblock \bibinfo{journal}{\emph{ArXiv}}  \bibinfo{volume}{abs/2404.13306}
  (\bibinfo{year}{2024}).
\newblock
\urldef\tempurl%
\url{https://api.semanticscholar.org/CorpusID:269293612}
\showURL{%
\tempurl}


\bibitem[Liao et~al\mbox{.}(2024)]%
        {QSpatial}
\bibfield{author}{\bibinfo{person}{Yuan-Hong Liao}, \bibinfo{person}{Rafid
  Mahmood}, \bibinfo{person}{Sanja Fidler}, {and} \bibinfo{person}{David
  Acuna}.} \bibinfo{year}{2024}\natexlab{}.
\newblock \bibinfo{title}{Reasoning Paths with Reference Objects Elicit
  Quantitative Spatial Reasoning in Large Vision-Language Models}.
\newblock
\showeprint[arxiv]{2409.09788}~[cs.CV]
\urldef\tempurl%
\url{https://arxiv.org/abs/2409.09788}
\showURL{%
\tempurl}


\bibitem[Lin et~al\mbox{.}(2024)]%
        {VILA}
\bibfield{author}{\bibinfo{person}{Ji Lin}, \bibinfo{person}{Hongxu Yin},
  \bibinfo{person}{Wei Ping}, \bibinfo{person}{Pavlo Molchanov},
  \bibinfo{person}{Mohammad Shoeybi}, {and} \bibinfo{person}{Song Han}.}
  \bibinfo{year}{2024}\natexlab{}.
\newblock \showarticletitle{VILA: On Pre-training for Visual Language Models}.
  In \bibinfo{booktitle}{\emph{Proceedings of the IEEE/CVF Conference on
  Computer Vision and Pattern Recognition (CVPR)}}.
  \bibinfo{pages}{26689--26699}.
\newblock


\bibitem[Liu et~al\mbox{.}(2023a)]%
        {ImprovedLLaVA}
\bibfield{author}{\bibinfo{person}{Haotian Liu}, \bibinfo{person}{Chunyuan Li},
  \bibinfo{person}{Yuheng Li}, {and} \bibinfo{person}{Yong~Jae Lee}.}
  \bibinfo{year}{2023}\natexlab{a}.
\newblock \bibinfo{title}{Improved Baselines with Visual Instruction Tuning}.
\newblock


\bibitem[Liu et~al\mbox{.}(2024)]%
        {LLaVANeXT}
\bibfield{author}{\bibinfo{person}{Haotian Liu}, \bibinfo{person}{Chunyuan Li},
  \bibinfo{person}{Yuheng Li}, \bibinfo{person}{Bo Li},
  \bibinfo{person}{Yuanhan Zhang}, \bibinfo{person}{Sheng Shen}, {and}
  \bibinfo{person}{Yong~Jae Lee}.} \bibinfo{year}{2024}\natexlab{}.
\newblock \bibinfo{title}{LLaVA-NeXT: Improved reasoning, OCR, and world
  knowledge}.
\newblock
\urldef\tempurl%
\url{https://llava-vl.github.io/blog/2024-01-30-llava-next/}
\showURL{%
\tempurl}


\bibitem[Liu et~al\mbox{.}(2023b)]%
        {LLaVA}
\bibfield{author}{\bibinfo{person}{Haotian Liu}, \bibinfo{person}{Chunyuan Li},
  \bibinfo{person}{Qingyang Wu}, {and} \bibinfo{person}{Yong~Jae Lee}.}
  \bibinfo{year}{2023}\natexlab{b}.
\newblock \bibinfo{title}{Visual Instruction Tuning}.
\newblock


\bibitem[Liu et~al\mbox{.}(2020)]%
        {ArtifactTextureDefake}
\bibfield{author}{\bibinfo{person}{Zhengzhe Liu}, \bibinfo{person}{Xiaojuan
  Qi}, {and} \bibinfo{person}{Philip~H.S. Torr}.}
  \bibinfo{year}{2020}\natexlab{}.
\newblock \showarticletitle{Global Texture Enhancement for Fake Face Detection
  in the Wild}. In \bibinfo{booktitle}{\emph{2020 IEEE/CVF Conference on
  Computer Vision and Pattern Recognition (CVPR)}}.
  \bibinfo{pages}{8057--8066}.
\newblock
\href{https://doi.org/10.1109/CVPR42600.2020.00808}{doi:\nolinkurl{10.1109/CVPR42600.2020.00808}}


\bibitem[Lorenz et~al\mbox{.}(2023)]%
        {IntrinsicDimensionalities}
\bibfield{author}{\bibinfo{person}{Peter Lorenz}, \bibinfo{person}{Ricard
  Durall}, {and} \bibinfo{person}{Janis Keuper}.}
  \bibinfo{year}{2023}\natexlab{}.
\newblock \showarticletitle{Detecting Images Generated by Deep Diffusion Models
  using their Local Intrinsic Dimensionality}.
\newblock \bibinfo{journal}{\emph{2023 IEEE/CVF International Conference on
  Computer Vision Workshops (ICCVW)}} (\bibinfo{year}{2023}),
  \bibinfo{pages}{448--459}.
\newblock
\urldef\tempurl%
\url{https://api.semanticscholar.org/CorpusID:259342331}
\showURL{%
\tempurl}


\bibitem[Lu et~al\mbox{.}(2024a)]%
        {MathVista}
\bibfield{author}{\bibinfo{person}{Pan Lu}, \bibinfo{person}{Hritik Bansal},
  \bibinfo{person}{Tony Xia}, \bibinfo{person}{Jiacheng Liu},
  \bibinfo{person}{Chunyuan Li}, \bibinfo{person}{Hannaneh Hajishirzi},
  \bibinfo{person}{Hao Cheng}, \bibinfo{person}{Kai-Wei Chang},
  \bibinfo{person}{Michel Galley}, {and} \bibinfo{person}{Jianfeng Gao}.}
  \bibinfo{year}{2024}\natexlab{a}.
\newblock \showarticletitle{MathVista: Evaluating Mathematical Reasoning of
  Foundation Models in Visual Contexts}. In
  \bibinfo{booktitle}{\emph{International Conference on Learning
  Representations (ICLR)}}.
\newblock


\bibitem[Lu et~al\mbox{.}(2024b)]%
        {Ovis}
\bibfield{author}{\bibinfo{person}{Shiyin Lu}, \bibinfo{person}{Yang Li},
  \bibinfo{person}{Qing-Guo Chen}, \bibinfo{person}{Zhao Xu},
  \bibinfo{person}{Weihua Luo}, \bibinfo{person}{Kaifu Zhang}, {and}
  \bibinfo{person}{Han-Jia Ye}.} \bibinfo{year}{2024}\natexlab{b}.
\newblock \showarticletitle{Ovis: Structural Embedding Alignment for Multimodal
  Large Language Model}.
\newblock \bibinfo{journal}{\emph{arXiv:2405.20797}} (\bibinfo{year}{2024}).
\newblock


\bibitem[Ojha et~al\mbox{.}(2023)]%
        {NPR}
\bibfield{author}{\bibinfo{person}{Utkarsh Ojha}, \bibinfo{person}{Yuheng Li},
  {and} \bibinfo{person}{Yong~Jae Lee}.} \bibinfo{year}{2023}\natexlab{}.
\newblock \showarticletitle{Towards Universal Fake Image Detectors that
  Generalize Across Generative Models}.
\newblock \bibinfo{journal}{\emph{2023 IEEE/CVF Conference on Computer Vision
  and Pattern Recognition (CVPR)}} (\bibinfo{year}{2023}),
  \bibinfo{pages}{24480--24489}.
\newblock
\urldef\tempurl%
\url{https://api.semanticscholar.org/CorpusID:257038440}
\showURL{%
\tempurl}


\bibitem[OpenAI(2023)]%
        {DALLE3}
\bibfield{author}{\bibinfo{person}{OpenAI}.} \bibinfo{year}{2023}\natexlab{}.
\newblock \bibinfo{booktitle}{\emph{DALL·E 3 System Card}}.
\newblock
\urldef\tempurl%
\url{https://cdn.openai.com/papers/DALL_E_3_System_Card.pdf}
\showURL{%
\tempurl}


\bibitem[{OpenAI}(2024a)]%
        {GPT4oMini}
\bibfield{author}{\bibinfo{person}{{OpenAI}}.}
  \bibinfo{year}{2024}\natexlab{a}.
\newblock \bibinfo{title}{GPT-4o mini: advancing cost-efficient intelligence}.
\newblock
  \bibinfo{howpublished}{https://openai.com/index/gpt-4o-mini-advancing-cost-efficient-intelligence/}.
\newblock


\bibitem[{OpenAI}(2024b)]%
        {GPT4o}
\bibfield{author}{\bibinfo{person}{{OpenAI}}.}
  \bibinfo{year}{2024}\natexlab{b}.
\newblock \bibinfo{title}{GPT-4o System Card}.
\newblock
\showeprint[arxiv]{2410.21276}~[cs.CL]
\urldef\tempurl%
\url{https://arxiv.org/abs/2410.21276}
\showURL{%
\tempurl}


\bibitem[Oquab et~al\mbox{.}(2024)]%
        {DinoV2}
\bibfield{author}{\bibinfo{person}{Maxime Oquab}, \bibinfo{person}{Timothée
  Darcet}, \bibinfo{person}{Théo Moutakanni}, \bibinfo{person}{Huy Vo},
  \bibinfo{person}{Marc Szafraniec}, \bibinfo{person}{Vasil Khalidov},
  \bibinfo{person}{Pierre Fernandez}, \bibinfo{person}{Daniel Haziza},
  \bibinfo{person}{Francisco Massa}, \bibinfo{person}{Alaaeldin El-Nouby},
  \bibinfo{person}{Mahmoud Assran}, \bibinfo{person}{Nicolas Ballas},
  \bibinfo{person}{Wojciech Galuba}, \bibinfo{person}{Russell Howes},
  \bibinfo{person}{Po-Yao Huang}, \bibinfo{person}{Shang-Wen Li},
  \bibinfo{person}{Ishan Misra}, \bibinfo{person}{Michael Rabbat},
  \bibinfo{person}{Vasu Sharma}, \bibinfo{person}{Gabriel Synnaeve},
  \bibinfo{person}{Hu Xu}, \bibinfo{person}{Hervé Jegou},
  \bibinfo{person}{Julien Mairal}, \bibinfo{person}{Patrick Labatut},
  \bibinfo{person}{Armand Joulin}, {and} \bibinfo{person}{Piotr Bojanowski}.}
  \bibinfo{year}{2024}\natexlab{}.
\newblock \bibinfo{title}{DINOv2: Learning Robust Visual Features without
  Supervision}.
\newblock
\showeprint[arxiv]{2304.07193}~[cs.CV]
\urldef\tempurl%
\url{https://arxiv.org/abs/2304.07193}
\showURL{%
\tempurl}


\bibitem[Park and Owens(2024)]%
        {ComFor}
\bibfield{author}{\bibinfo{person}{Jeongsoo Park} {and} \bibinfo{person}{Andrew
  Owens}.} \bibinfo{year}{2024}\natexlab{}.
\newblock \bibinfo{title}{Community Forensics: Using Thousands of Generators to
  Train Fake Image Detectors}.
\newblock
\showeprint[arxiv]{2411.04125}~[cs.CV]
\urldef\tempurl%
\url{https://arxiv.org/abs/2411.04125}
\showURL{%
\tempurl}


\bibitem[{Qwen Team}(2024)]%
        {QVQ}
\bibfield{author}{\bibinfo{person}{{Qwen Team}}.}
  \bibinfo{year}{2024}\natexlab{}.
\newblock \bibinfo{title}{QVQ: To See the World with Wisdom}.
\newblock
\urldef\tempurl%
\url{https://qwenlm.github.io/blog/qvq-72b-preview/}
\showURL{%
\tempurl}


\bibitem[{Qwen Team}(2025)]%
        {Qwen25VL}
\bibfield{author}{\bibinfo{person}{{Qwen Team}}.}
  \bibinfo{year}{2025}\natexlab{}.
\newblock \bibinfo{title}{Qwen2.5-VL}.
\newblock
\urldef\tempurl%
\url{https://qwenlm.github.io/blog/qwen2.5-vl/}
\showURL{%
\tempurl}


\bibitem[Ricker et~al\mbox{.}(2024)]%
        {AEROBLADE}
\bibfield{author}{\bibinfo{person}{Jonas Ricker}, \bibinfo{person}{Denis
  Lukovnikov}, {and} \bibinfo{person}{Asja Fischer}.}
  \bibinfo{year}{2024}\natexlab{}.
\newblock \showarticletitle{AEROBLADE: Training-Free Detection of Latent
  Diffusion Images Using Autoencoder Reconstruction Error}.
\newblock \bibinfo{journal}{\emph{2024 IEEE/CVF Conference on Computer Vision
  and Pattern Recognition (CVPR)}} (\bibinfo{year}{2024}),
  \bibinfo{pages}{9130--9140}.
\newblock
\urldef\tempurl%
\url{https://api.semanticscholar.org/CorpusID:267335007}
\showURL{%
\tempurl}


\bibitem[Rombach et~al\mbox{.}(2022)]%
        {SD21}
\bibfield{author}{\bibinfo{person}{Robin Rombach}, \bibinfo{person}{Andreas
  Blattmann}, \bibinfo{person}{Dominik Lorenz}, \bibinfo{person}{Patrick
  Esser}, {and} \bibinfo{person}{Bj\"orn Ommer}.}
  \bibinfo{year}{2022}\natexlab{}.
\newblock \bibinfo{title}{High-Resolution Image Synthesis With Latent Diffusion
  Models}.
\newblock \bibinfo{numpages}{10684-10695}~pages.
\newblock


\bibitem[Saharia et~al\mbox{.}(2022)]%
        {Imagen}
\bibfield{author}{\bibinfo{person}{Chitwan Saharia}, \bibinfo{person}{William
  Chan}, \bibinfo{person}{Saurabh Saxena}, \bibinfo{person}{Lala Li},
  \bibinfo{person}{Jay Whang}, \bibinfo{person}{Emily~L Denton},
  \bibinfo{person}{Kamyar Ghasemipour}, \bibinfo{person}{Raphael
  Gontijo~Lopes}, \bibinfo{person}{Burcu Karagol~Ayan}, \bibinfo{person}{Tim
  Salimans}, {et~al\mbox{.}}} \bibinfo{year}{2022}\natexlab{}.
\newblock \showarticletitle{Photorealistic text-to-image diffusion models with
  deep language understanding}.
\newblock \bibinfo{journal}{\emph{Advances in Neural Information Processing
  Systems}}  \bibinfo{volume}{35} (\bibinfo{year}{2022}),
  \bibinfo{pages}{36479--36494}.
\newblock


\bibitem[Schuhmann et~al\mbox{.}(2022)]%
        {LAION5B}
\bibfield{author}{\bibinfo{person}{Christoph Schuhmann},
  \bibinfo{person}{Romain Beaumont}, \bibinfo{person}{Richard Vencu},
  \bibinfo{person}{Cade Gordon}, \bibinfo{person}{Ross Wightman},
  \bibinfo{person}{Mehdi Cherti}, \bibinfo{person}{Theo Coombes},
  \bibinfo{person}{Aarush Katta}, \bibinfo{person}{Clayton Mullis},
  \bibinfo{person}{Mitchell Wortsman}, {et~al\mbox{.}}}
  \bibinfo{year}{2022}\natexlab{}.
\newblock \showarticletitle{Laion-5b: An open large-scale dataset for training
  next generation image-text models}.
\newblock \bibinfo{journal}{\emph{Advances in Neural Information Processing
  Systems}}  \bibinfo{volume}{35} (\bibinfo{year}{2022}),
  \bibinfo{pages}{25278--25294}.
\newblock


\bibitem[Selvaraju et~al\mbox{.}(2017)]%
        {XplainedGradCAM}
\bibfield{author}{\bibinfo{person}{Ramprasaath~R. Selvaraju},
  \bibinfo{person}{Michael Cogswell}, \bibinfo{person}{Abhishek Das},
  \bibinfo{person}{Ramakrishna Vedantam}, \bibinfo{person}{Devi Parikh}, {and}
  \bibinfo{person}{Dhruv Batra}.} \bibinfo{year}{2017}\natexlab{}.
\newblock \showarticletitle{Grad-CAM: Visual Explanations from Deep Networks
  via Gradient-Based Localization}. In \bibinfo{booktitle}{\emph{2017 IEEE
  International Conference on Computer Vision (ICCV)}}.
  \bibinfo{pages}{618--626}.
\newblock
\href{https://doi.org/10.1109/ICCV.2017.74}{doi:\nolinkurl{10.1109/ICCV.2017.74}}


\bibitem[Simonyan et~al\mbox{.}(2013)]%
        {XplainedCNN}
\bibfield{author}{\bibinfo{person}{Karen Simonyan}, \bibinfo{person}{Andrea
  Vedaldi}, {and} \bibinfo{person}{Andrew Zisserman}.}
  \bibinfo{year}{2013}\natexlab{}.
\newblock \showarticletitle{Deep Inside Convolutional Networks: Visualising
  Image Classification Models and Saliency Maps}.
\newblock \bibinfo{journal}{\emph{CoRR}}  \bibinfo{volume}{abs/1312.6034}
  (\bibinfo{year}{2013}).
\newblock
\urldef\tempurl%
\url{https://api.semanticscholar.org/CorpusID:1450294}
\showURL{%
\tempurl}


\bibitem[Song et~al\mbox{.}(2020)]%
        {DDIM}
\bibfield{author}{\bibinfo{person}{Jiaming Song}, \bibinfo{person}{Chenlin
  Meng}, {and} \bibinfo{person}{Stefano Ermon}.}
  \bibinfo{year}{2020}\natexlab{}.
\newblock \showarticletitle{Denoising Diffusion Implicit Models}.
\newblock \bibinfo{journal}{\emph{ArXiv}}  \bibinfo{volume}{abs/2010.02502}
  (\bibinfo{year}{2020}).
\newblock
\urldef\tempurl%
\url{https://api.semanticscholar.org/CorpusID:222140788}
\showURL{%
\tempurl}


\bibitem[Sundararajan et~al\mbox{.}(2017)]%
        {XplainedAxiomaticAttr}
\bibfield{author}{\bibinfo{person}{Mukund Sundararajan}, \bibinfo{person}{Ankur
  Taly}, {and} \bibinfo{person}{Qiqi Yan}.} \bibinfo{year}{2017}\natexlab{}.
\newblock \showarticletitle{Axiomatic attribution for deep networks}. In
  \bibinfo{booktitle}{\emph{Proceedings of the 34th International Conference on
  Machine Learning - Volume 70}} (Sydney, NSW, Australia)
  \emph{(\bibinfo{series}{ICML'17})}. \bibinfo{publisher}{JMLR.org},
  \bibinfo{pages}{3319–3328}.
\newblock


\bibitem[Tan and Le(2019)]%
        {EfficientNet}
\bibfield{author}{\bibinfo{person}{Mingxing Tan} {and} \bibinfo{person}{Quoc~V.
  Le}.} \bibinfo{year}{2019}\natexlab{}.
\newblock \showarticletitle{EfficientNet: Rethinking Model Scaling for
  Convolutional Neural Networks}.
\newblock \bibinfo{journal}{\emph{ArXiv}}  \bibinfo{volume}{abs/1905.11946}
  (\bibinfo{year}{2019}).
\newblock
\urldef\tempurl%
\url{https://api.semanticscholar.org/CorpusID:167217261}
\showURL{%
\tempurl}


\bibitem[Tao et~al\mbox{.}(2023)]%
        {GALIP}
\bibfield{author}{\bibinfo{person}{Ming Tao}, \bibinfo{person}{Bing-Kun Bao},
  \bibinfo{person}{Hao Tang}, {and} \bibinfo{person}{Changsheng Xu}.}
  \bibinfo{year}{2023}\natexlab{}.
\newblock \showarticletitle{GALIP: Generative Adversarial CLIPs for
  Text-to-Image Synthesis}. In \bibinfo{booktitle}{\emph{Proceedings of the
  IEEE/CVF Conference on Computer Vision and Pattern Recognition}}.
  \bibinfo{pages}{14214--14223}.
\newblock


\bibitem[Tao et~al\mbox{.}(2022)]%
        {DFGAN}
\bibfield{author}{\bibinfo{person}{Ming Tao}, \bibinfo{person}{Hao Tang},
  \bibinfo{person}{Fei Wu}, \bibinfo{person}{Xiao-Yuan Jing},
  \bibinfo{person}{Bing-Kun Bao}, {and} \bibinfo{person}{Changsheng Xu}.}
  \bibinfo{year}{2022}\natexlab{}.
\newblock \showarticletitle{Df-gan: A simple and effective baseline for
  text-to-image synthesis}. In \bibinfo{booktitle}{\emph{Proceedings of the
  IEEE/CVF Conference on Computer Vision and Pattern Recognition}}.
  \bibinfo{pages}{16515--16525}.
\newblock


\bibitem[{The Llama Team}(2024)]%
        {Llama3}
\bibfield{author}{\bibinfo{person}{{The Llama Team}}.}
  \bibinfo{year}{2024}\natexlab{}.
\newblock \bibinfo{title}{The Llama 3 Herd of Models}.
\newblock
\showeprint[arxiv]{2407.21783}~[cs.AI]
\urldef\tempurl%
\url{https://arxiv.org/abs/2407.21783}
\showURL{%
\tempurl}


\bibitem[Van Den~Oord et~al\mbox{.}(2017)]%
        {VQVAE}
\bibfield{author}{\bibinfo{person}{Aaron Van Den~Oord}, \bibinfo{person}{Oriol
  Vinyals}, {et~al\mbox{.}}} \bibinfo{year}{2017}\natexlab{}.
\newblock \showarticletitle{Neural discrete representation learning}.
\newblock \bibinfo{journal}{\emph{Advances in neural information processing
  systems}}  \bibinfo{volume}{30} (\bibinfo{year}{2017}).
\newblock


\bibitem[Wang et~al\mbox{.}(2022)]%
        {ObjectFormer}
\bibfield{author}{\bibinfo{person}{Junke Wang}, \bibinfo{person}{Zuxuan Wu},
  \bibinfo{person}{Jingjing Chen}, \bibinfo{person}{Xintong Han},
  \bibinfo{person}{Abhinav Shrivastava}, \bibinfo{person}{Ser-Nam Lim}, {and}
  \bibinfo{person}{Yu-Gang Jiang}.} \bibinfo{year}{2022}\natexlab{}.
\newblock \showarticletitle{Objectformer for image manipulation detection and
  localization}. In \bibinfo{booktitle}{\emph{Proceedings of the IEEE/CVF
  Conference on Computer Vision and Pattern Recognition}}.
\newblock


\bibitem[Wang et~al\mbox{.}(2024)]%
        {Qwen2VL}
\bibfield{author}{\bibinfo{person}{Peng Wang}, \bibinfo{person}{Shuai Bai},
  \bibinfo{person}{Sinan Tan}, \bibinfo{person}{Shijie Wang},
  \bibinfo{person}{Zhihao Fan}, \bibinfo{person}{Jinze Bai},
  \bibinfo{person}{Keqin Chen}, \bibinfo{person}{Xuejing Liu},
  \bibinfo{person}{Jialin Wang}, \bibinfo{person}{Wenbin Ge},
  \bibinfo{person}{Yang Fan}, \bibinfo{person}{Kai Dang},
  \bibinfo{person}{Mengfei Du}, \bibinfo{person}{Xuancheng Ren},
  \bibinfo{person}{Rui Men}, \bibinfo{person}{Dayiheng Liu},
  \bibinfo{person}{Chang Zhou}, \bibinfo{person}{Jingren Zhou}, {and}
  \bibinfo{person}{Junyang Lin}.} \bibinfo{year}{2024}\natexlab{}.
\newblock \showarticletitle{Qwen2-VL: Enhancing Vision-Language Model's
  Perception of the World at Any Resolution}.
\newblock \bibinfo{journal}{\emph{arXiv preprint arXiv:2409.12191}}
  (\bibinfo{year}{2024}).
\newblock


\bibitem[Wang et~al\mbox{.}(2020)]%
        {CNNSpot}
\bibfield{author}{\bibinfo{person}{Sheng-Yu Wang}, \bibinfo{person}{Oliver
  Wang}, \bibinfo{person}{Richard Zhang}, \bibinfo{person}{Andrew Owens}, {and}
  \bibinfo{person}{Alexei~A Efros}.} \bibinfo{year}{2020}\natexlab{}.
\newblock \showarticletitle{CNN-generated images are surprisingly easy to
  spot... for now}. In \bibinfo{booktitle}{\emph{Proceedings of the IEEE/CVF
  conference on computer vision and pattern recognition}}.
  \bibinfo{pages}{8695--8704}.
\newblock


\bibitem[Wu et~al\mbox{.}(2024)]%
        {VLMPrompting}
\bibfield{author}{\bibinfo{person}{Junda Wu}, \bibinfo{person}{Zhehao Zhang},
  \bibinfo{person}{Yu Xia}, \bibinfo{person}{Xintong Li},
  \bibinfo{person}{Zhaoyang Xia}, \bibinfo{person}{Aaron Chang},
  \bibinfo{person}{Tong Yu}, \bibinfo{person}{Sungchul Kim},
  \bibinfo{person}{Ryan~A. Rossi}, \bibinfo{person}{Ruiyi Zhang},
  \bibinfo{person}{Subrata Mitra}, \bibinfo{person}{Dimitris~N. Metaxas},
  \bibinfo{person}{Lina Yao}, \bibinfo{person}{Jingbo Shang}, {and}
  \bibinfo{person}{Julian McAuley}.} \bibinfo{year}{2024}\natexlab{}.
\newblock \bibinfo{title}{Visual Prompting in Multimodal Large Language Models:
  A Survey}.
\newblock
\showeprint[arxiv]{2409.15310}~[cs.LG]
\urldef\tempurl%
\url{https://arxiv.org/abs/2409.15310}
\showURL{%
\tempurl}


\bibitem[Wu and Xie(2023)]%
        {VStar}
\bibfield{author}{\bibinfo{person}{Penghao Wu} {and} \bibinfo{person}{Saining
  Xie}.} \bibinfo{year}{2023}\natexlab{}.
\newblock \bibinfo{title}{V*: Guided Visual Search as a Core Mechanism in
  Multimodal LLMs}.
\newblock
\showeprint[arxiv]{2312.14135}~[cs.CV]
\urldef\tempurl%
\url{https://arxiv.org/abs/2312.14135}
\showURL{%
\tempurl}


\bibitem[Xu et~al\mbox{.}(2024)]%
        {LLaVACoT}
\bibfield{author}{\bibinfo{person}{Guowei Xu}, \bibinfo{person}{Peng Jin},
  \bibinfo{person}{Hao Li}, \bibinfo{person}{Yibing Song},
  \bibinfo{person}{Lichao Sun}, {and} \bibinfo{person}{Li Yuan}.}
  \bibinfo{year}{2024}\natexlab{}.
\newblock \showarticletitle{LLaVA-CoT: Let Vision Language Models Reason
  Step-by-Step}.
\newblock \bibinfo{journal}{\emph{ArXiv}}  \bibinfo{volume}{abs/2411.10440}
  (\bibinfo{year}{2024}).
\newblock
\urldef\tempurl%
\url{https://api.semanticscholar.org/CorpusID:274116688}
\showURL{%
\tempurl}


\bibitem[Ye et~al\mbox{.}(2024)]%
        {MPlugOwl2}
\bibfield{author}{\bibinfo{person}{Qinghao Ye}, \bibinfo{person}{Haiyang Xu},
  \bibinfo{person}{Jiabo Ye}, \bibinfo{person}{Ming Yan},
  \bibinfo{person}{Anwen Hu}, \bibinfo{person}{Haowei Liu}, \bibinfo{person}{Qi
  Qian}, \bibinfo{person}{Ji Zhang}, {and} \bibinfo{person}{Fei Huang}.}
  \bibinfo{year}{2024}\natexlab{}.
\newblock \showarticletitle{mplug-owl2: Revolutionizing multi-modal large
  language model with modality collaboration}. In
  \bibinfo{booktitle}{\emph{Proceedings of the IEEE/CVF Conference on Computer
  Vision and Pattern Recognition}}. \bibinfo{pages}{13040--13051}.
\newblock


\bibitem[Yeh et~al\mbox{.}(2023)]%
        {SDLyCORIS}
\bibfield{author}{\bibinfo{person}{Shin-Ying Yeh}, \bibinfo{person}{Yu-Guan
  Hsieh}, \bibinfo{person}{Zhidong Gao}, \bibinfo{person}{Bernard~BW Yang},
  \bibinfo{person}{Giyeong Oh}, {and} \bibinfo{person}{Yanmin Gong}.}
  \bibinfo{year}{2023}\natexlab{}.
\newblock \showarticletitle{Navigating Text-To-Image Customization: From
  LyCORIS Fine-Tuning to Model Evaluation}.
\newblock \bibinfo{journal}{\emph{arXiv preprint arXiv:2309.14859}}
  (\bibinfo{year}{2023}).
\newblock


\bibitem[Ying et~al\mbox{.}(2024)]%
        {MMTBench}
\bibfield{author}{\bibinfo{person}{Kaining Ying}, \bibinfo{person}{Fanqing
  Meng}, \bibinfo{person}{Jin Wang}, \bibinfo{person}{Zhiqian Li},
  \bibinfo{person}{Han Lin}, \bibinfo{person}{Yue Yang}, \bibinfo{person}{Hao
  Zhang}, \bibinfo{person}{Wenbo Zhang}, \bibinfo{person}{Yuqi Lin},
  \bibinfo{person}{Shuo Liu}, \bibinfo{person}{Jiayi Lei},
  \bibinfo{person}{Quanfeng Lu}, \bibinfo{person}{Runjian Chen},
  \bibinfo{person}{Peng Xu}, \bibinfo{person}{Renrui Zhang},
  \bibinfo{person}{Haozhe Zhang}, \bibinfo{person}{Peng Gao},
  \bibinfo{person}{Yali Wang}, \bibinfo{person}{Yu Qiao}, \bibinfo{person}{Ping
  Luo}, \bibinfo{person}{Kaipeng Zhang}, {and} \bibinfo{person}{Wenqi Shao}.}
  \bibinfo{year}{2024}\natexlab{}.
\newblock \bibinfo{title}{MMT-Bench: A Comprehensive Multimodal Benchmark for
  Evaluating Large Vision-Language Models Towards Multitask AGI}.
\newblock
\showeprint[arxiv]{2404.16006}~[cs.CV]


\bibitem[Yu et~al\mbox{.}(2015)]%
        {LSUN}
\bibfield{author}{\bibinfo{person}{Fisher Yu}, \bibinfo{person}{Ari Seff},
  \bibinfo{person}{Yinda Zhang}, \bibinfo{person}{Shuran Song},
  \bibinfo{person}{Thomas Funkhouser}, {and} \bibinfo{person}{Jianxiong Xiao}.}
  \bibinfo{year}{2015}\natexlab{}.
\newblock \showarticletitle{Lsun: Construction of a large-scale image dataset
  using deep learning with humans in the loop}.
\newblock \bibinfo{journal}{\emph{arXiv preprint arXiv:1506.03365}}
  (\bibinfo{year}{2015}).
\newblock


\bibitem[Zhang et~al\mbox{.}(2024b)]%
        {DetailPerceptionInZeroShotVQA}
\bibfield{author}{\bibinfo{person}{Jiarui Zhang}, \bibinfo{person}{Mahyar
  Khayatkhoei}, \bibinfo{person}{Prateek Chhikara}, {and}
  \bibinfo{person}{Filip Ilievski}.} \bibinfo{year}{2024}\natexlab{b}.
\newblock \bibinfo{title}{Towards Perceiving Small Visual Details in Zero-shot
  Visual Question Answering with Multimodal LLMs}.
\newblock
\showeprint[arxiv]{2310.16033}~[cs.CV]
\urldef\tempurl%
\url{https://arxiv.org/abs/2310.16033}
\showURL{%
\tempurl}


\bibitem[Zhang et~al\mbox{.}(2023)]%
        {ControlNet}
\bibfield{author}{\bibinfo{person}{Lvmin Zhang}, \bibinfo{person}{Anyi Rao},
  {and} \bibinfo{person}{Maneesh Agrawala}.} \bibinfo{year}{2023}\natexlab{}.
\newblock \showarticletitle{Adding conditional control to text-to-image
  diffusion models}. In \bibinfo{booktitle}{\emph{Proceedings of the IEEE/CVF
  International Conference on Computer Vision}}. \bibinfo{pages}{3836--3847}.
\newblock


\bibitem[Zhang et~al\mbox{.}(2019)]%
        {ArtifactDetSimGAN}
\bibfield{author}{\bibinfo{person}{Xu Zhang}, \bibinfo{person}{Svebor Karaman},
  {and} \bibinfo{person}{Shih-Fu Chang}.} \bibinfo{year}{2019}\natexlab{}.
\newblock \showarticletitle{Detecting and Simulating Artifacts in GAN Fake
  Images}. In \bibinfo{booktitle}{\emph{2019 IEEE International Workshop on
  Information Forensics and Security (WIFS)}}. \bibinfo{pages}{1--6}.
\newblock
\href{https://doi.org/10.1109/WIFS47025.2019.9035107}{doi:\nolinkurl{10.1109/WIFS47025.2019.9035107}}


\bibitem[Zhang et~al\mbox{.}(2024a)]%
        {CommonSenseReasoning}
\bibfield{author}{\bibinfo{person}{Yue Zhang}, \bibinfo{person}{Ben Colman},
  \bibinfo{person}{Ali Shahriyari}, {and} \bibinfo{person}{Gaurav Bharaj}.}
  \bibinfo{year}{2024}\natexlab{a}.
\newblock \showarticletitle{Common Sense Reasoning for Deep Fake Detection}.
\newblock \bibinfo{journal}{\emph{ArXiv}}  \bibinfo{volume}{abs/2402.00126}
  (\bibinfo{year}{2024}).
\newblock
\urldef\tempurl%
\url{https://api.semanticscholar.org/CorpusID:267365533}
\showURL{%
\tempurl}


\bibitem[Zhang et~al\mbox{.}(2024c)]%
        {ABench}
\bibfield{author}{\bibinfo{person}{Zicheng Zhang}, \bibinfo{person}{Haoning
  Wu}, \bibinfo{person}{Chunyi Li}, \bibinfo{person}{Yingjie Zhou},
  \bibinfo{person}{Wei Sun}, \bibinfo{person}{Xiongkuo Min},
  \bibinfo{person}{Zijian Chen}, \bibinfo{person}{Xiaohong Liu},
  \bibinfo{person}{Weisi Lin}, {and} \bibinfo{person}{Guangtao Zhai}.}
  \bibinfo{year}{2024}\natexlab{c}.
\newblock \showarticletitle{A-Bench: Are LMMs Masters at Evaluating
  AI-generated Images?}
\newblock \bibinfo{journal}{\emph{ArXiv}}  \bibinfo{volume}{abs/2406.03070}
  (\bibinfo{year}{2024}).
\newblock
\urldef\tempurl%
\url{https://api.semanticscholar.org/CorpusID:270258348}
\showURL{%
\tempurl}


\bibitem[Zhao et~al\mbox{.}(2023)]%
        {MMIContextL}
\bibfield{author}{\bibinfo{person}{Haozhe Zhao}, \bibinfo{person}{Zefan Cai},
  \bibinfo{person}{Shuzheng Si}, \bibinfo{person}{Xiaojian Ma},
  \bibinfo{person}{Kaikai An}, \bibinfo{person}{Liang Chen},
  \bibinfo{person}{Zixuan Liu}, \bibinfo{person}{Sheng Wang},
  \bibinfo{person}{Wenjuan Han}, {and} \bibinfo{person}{Baobao Chang}.}
  \bibinfo{year}{2023}\natexlab{}.
\newblock \showarticletitle{MMICL: Empowering Vision-language Model with
  Multi-Modal In-Context Learning}.
\newblock \bibinfo{journal}{\emph{ArXiv}}  \bibinfo{volume}{abs/2309.07915}
  (\bibinfo{year}{2023}).
\newblock
\urldef\tempurl%
\url{https://api.semanticscholar.org/CorpusID:261823391}
\showURL{%
\tempurl}


\end{thebibliography}
